\documentclass[10pt,twocolumn,letterpaper]{article}

\usepackage[pagenumbers]{cvpr}            
\usepackage[utf8]{inputenc}
\usepackage{algorithm}
\usepackage{algpseudocode}
\usepackage[accsupp]{axessibility}
\usepackage{amssymb}
\usepackage{amsmath}    
\usepackage{bm}         
\usepackage{caption}    
\usepackage{booktabs}
\usepackage{multirow}
\usepackage[dvipsnames]{xcolor}
\usepackage{threeparttable}
\usepackage[table]{xcolor}
\usepackage{xcolor}
\usepackage{colortbl}
\definecolor{LavenderBlue}{RGB}{120,130,255}
\definecolor{SoftBlue}{RGB}{63,127,191}
\definecolor{cvprblue}{rgb}{0.21,0.49,0.74}
\usepackage[pagebackref,breaklinks,colorlinks,allcolors=cvprblue]{hyperref}

\def\paperID{12716} 
\def\confName{CVPR}
\def\confYear{2026}

\title{Mitigating Object Hallucination in LVLMs via Attention Imbalance Rectification}

\author{
 Han Sun\textsuperscript{1},
 Qin Li\textsuperscript{1},
 Peixin Wang\textsuperscript{1},
 Min Zhang\textsuperscript{1}\thanks{Corresponding author}
\\
 \textsuperscript{1}Shanghai Key Laboratory of Trustworthy Computing, East China Normal University \\
 \small{
  \texttt{hsun@stu.ecnu.edu.cn} \quad \texttt{\{qli, pxwang, zhangmin\}@sei.ecnu.edu.cn}
}
}

\begin{document}
\maketitle
\begin{abstract}
	
	Object hallucination in Large Vision-Language Models (LVLMs) severely compromises their reliability in real-world applications, posing a critical barrier to their deployment in high-stakes scenarios such as autonomous driving and medical image analysis.  
    Through systematic empirical investigation, we identify that the imbalanced attention allocation, both \textbf{across modalities} (i.e., vision and language) and \textbf{within modalities} (among individual tokens), exhibits a strong causal correlation with the occurrence of object hallucination.
	Leveraging this insight, we introduce a novel concept termed \emph{attention imbalance}, which not only quantifies the degree of attention disparity but also visually delineates the underlying patterns (e.g., over-attentiveness to irrelevant language tokens or under-attentiveness to discriminative visual features) that drive object hallucination. 
    To mitigate object hallucination, we further propose \textbf{A}ttention \textbf{I}mbalance \textbf{R}ectification (\textbf{AIR}), a lightweight decoding-time intervention method that reallocates attention weights and adjusts attention distributions to rectify modality-wise and token-wise imbalances. 
	Extensive evaluations on four mainstream LVLMs and three benchmarks (CHAIR, POPE, and MM-Vet) with seven baselines demonstrate that AIR consistently reduces object hallucination rates, achieving up to a 35.1\% reduction compared to the baselines, while improving up to 15.9\% of LVLMs' general capability across diverse vision-language tasks. The code is publicly available at \url{https://github.com/Ice-wave/AIR}.

\end{abstract}

\section{Introduction}
\label{sec:intro}

Large Vision-Language Models (LVLMs), with their powerful capability for unified cross-modal understanding and reasoning \cite{openai2024gpt4technicalreport, geminiteam2025geminifamilyhighlycapable, deepseekai2025deepseekv3technicalreport, sun2025vlmsdetectlocalizefinegrained}, have become essential tools for tasks ranging from fundamental perception such as image captioning \cite{10655294, 10.1007/978-3-031-73004-7_2, Lin2024VideoLLaVA, bao2022beit} to complex video reasoning \cite{10658097, chen2024mecd, 11093988, zhang2025iforeverevaluatingrealtime}. They are widely applied in critical industry domains such as autonomous driving \cite{sym16091196, chen2024multiobject, 10943406}, medical imaging \cite{https://doi.org/10.1002/aisy.202500255, chen2024detectingevaluatingmedicalhallucinations, yan2024medhvl}, and e-commerce \cite{li-etal-2024-multimodal, Khandelwal2023LargeSG}, where they substantially improve task efficiency and accuracy. However, the issue of \textit{object hallucination}, i.e., the generation of non-existent objects in visual inputs, remains a major challenge for researchers and significantly compromises the reliability and trustworthiness of LVLMs \cite{liu2024surveyhallucinationlargevisionlanguage, Rohrbach2018ObjectHI, li2023evaluating, PEFTGuard2025}.

\begin{figure}[t]
	\centering
	\includegraphics[width=0.476
	\textwidth]{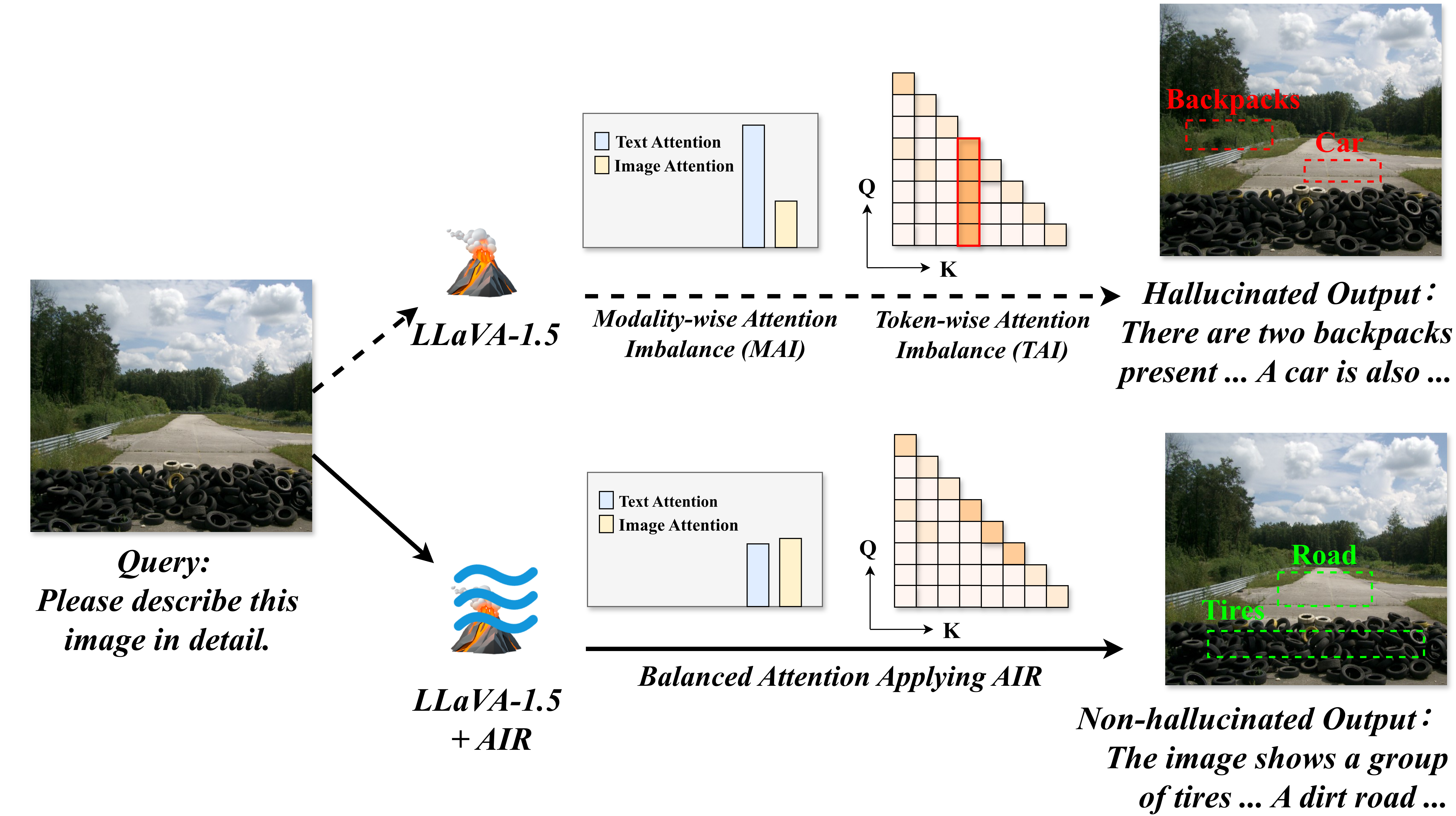}
	\caption{Attention imbalance causes LLaVA-1.5 to hallucinate (up) while hallucination disappears once the attention is balanced by our AIR approach (down). }
	\label{intro}
\end{figure}

Recent efforts to mitigating object hallucination in LVLMs can be broadly categorized into three types: \textbf{(1)} visual instruction fine-tuning and alignment \cite{Jiang_2024_CVPR, DBLP:conf/cvpr/Yu0WPYQT0Z24, 10656676}, \textbf{(2)} post-hoc techniques \cite{DBLP:conf/acl/FengS00BT24, wu-etal-2024-logical, yin2024woodpecker}, and \textbf{(3)} contrastive decoding \cite{10655750, DBLP:conf/cvpr/HuangDZ0H0L0Y24, Kim2023InstructiveDI, 10657718, sun2025syndecsynthesizethendecodeapproacharbitrary}.
The first category improves vision–language alignment through fine-grained supervision and preference optimization, achieving notable performance gains but at the expense of substantial training and data costs \cite{jing2025fgaif, varma2023villa}.
Post-hoc techniques detect, correct, or constrain hallucinations in model outputs through external mechanisms or auxiliary modules, but they introduce additional inference overhead \cite{li2025statisticalfactualityguaranteelarge, cekinel-etal-2025-multimodal}.
Contrastive decoding alleviates linguistic priors without retraining, providing a lightweight alternative, though its stability and generalizability remain limited \cite{wang-etal-2024-mitigating, chuang2024dola}.
In addition, identifying the root causes of hallucination remains a major challenge. This difficulty primarily stems from the complex training pipelines and underlying architectures of LVLMs, which hinder interpretability analysis. Although several studies have explored object hallucination from an interpretability perspective by examining factors such as visual information interaction \cite{10657718, DBLP:conf/iclr/YangL0X25, tang2025seeing}, positional encoding \cite{NEURIPS2024_a76ed4a8, zhaoMCA}, and anomalous tokens \cite{DBLP:conf/cvpr/HuangDZ0H0L0Y24, tang2025intervening}, a comprehensive and in-depth understanding is still lacking.

In this paper, we introduce a novel concept termed \textit{attention imbalance} (detailed in Section~\ref{4}) encompassing both modality-wise and token-wise imbalance, which serves to quantify the extent of attention disproportion across modalities and tokens.
Leveraging \textit{attention imbalance}, we investigate its role in driving object hallucination in LVLMs.
First, by analyzing object hallucination in the outputs of LVLMs, we observe a significant co-occurrence between token-wise attention imbalance and object hallucination.
We then refine our analysis to the attention-head level. 
By examining attention heads that are particularly prone to producing hallucinations, average heads, and those that rarely contribute to hallucinations, we observe that the hallucination-prone ones exhibit a markedly stronger modality-wise attention imbalance than the others.
Our further analysis uncovers the underlying reason for this discrepancy: these hallucination-prone heads closely inherit the attention patterns of their base language models.

Based on the analysis of attention imbalance, we propose \textbf{A}ttention \textbf{I}mbalance \textbf{R}ectification (\textbf{AIR}) (detailed in Section~\ref{5}) to correct overly imbalanced attention and mitigate hallucination.
\textbf{AIR} operates on two dimensions:
\textbf{(1)} \textit{Modality-Balanced Attention Reallocation}, which reallocates attention to alleviate excessive attention imbalance across modalities, and
\textbf{(2)} \textit{Variance-Constrained Projection Regularization}, which constrains the variance of the attention matrix to homogenize attention distributions. 
Figure \ref{intro} illustrates two types of attention imbalance and shows that \textbf{AIR}  mitigates object hallucination by rectifying attention imbalance in LVLMs.
Experiments on both hallucination and general LVLM benchmarks demonstrate that \textbf{AIR}  achieves superior hallucination mitigation while improving the model’s overall capabilities on general tasks.

The key contributions can be summarized as follows:
\begin{enumerate}
	
	\item We introduce a novel concept named  \textit{attention imbalance} to explain object hallucinations in LVLMs. \textit{Attention imbalance} not only quantifies the degree of attention disparity across two dimensions but also visually reveals underlying patterns, such as over-attention to irrelevant language tokens or under-attention to discriminative visual features, that drive object hallucination.

	\item 
	We propose \textbf{A}ttention \textbf{I}mbalance \textbf{R}ectification (\textbf{AIR}), a lightweight decoding-time intervention method that effectively mitigates hallucination through head-specific attention reallocation and homogenization, without incurring any additional training costs.
	
	\item Extensive experiments on four representative LVLMs demonstrate that \textbf{AIR} achieves superior performance in reducing hallucinations, decreasing the sentence-level and image-level hallucination rates $C_S, C_I$ by up to \textbf{35.1\%} and \textbf{22.6\%}, respectively, while  improving up to \textbf{15.9\%} of the general capability of LVLMs. 

\end{enumerate}

\section{Related Work}
\noindent\textbf{LVLMs.}
Large vision-language models are emerging multimodal models that jointly process visual and textual information, offering strong   representational and generalization abilities in complex vision–language tasks \cite{cai2024vipllava, hu2024matryoshka, 10.1007/978-3-031-73004-7_2, Lin2024VideoLLaVA, bao2022beit, pmlr}. Their architectures typically consist of three components: \textbf{(1)} a vision branch to encode images (e.g., CLIP \cite{Radford2021LearningTV} or EVA \cite{Fang2023EVA}); \textbf{(2)} a modality connector to align visual features with text embeddings; and \textbf{(3)} a pretrained language model to generate responses from both image and text inputs.
In practice, LVLMs adopt different strategies for modality alignment. For instance, LLaVA-1.5 \cite{10655294} employs a multilayer perceptron (MLP) to map the outputs of the vision branch into 576 image embeddings, Shikra \cite{Chen2023ShikraUM} applies a linear layer for feature alignment, and MiniGPT-4 \cite{zhu2024minigpt} leverages a learnable querying transformer that establishes vision–language connections using only 32 image tokens. Despite strong performance on multimodal reasoning tasks, these models still suffer from object hallucination, which undermines their reliability and applicability.

\noindent\textbf{Object Hallucinations in LVLMs.}
Object hallucination in LVLMs refers to generating descriptions of non-existent objects in an image. In high-stakes domains such as autonomous driving \cite{sym16091196, chen2024multiobject, 10943406}, UAV perception \cite{10943891, liu2024surveyhallucinationlargevisionlanguage} and medical imaging \cite{https://doi.org/10.1002/aisy.202500255, chen2024detectingevaluatingmedicalhallucinations, yan2024medhvl, zhang2024radflagblackboxhallucinationdetection}, object hallucination can distort perception, leading to serious safety risks. Recent research on mitigating hallucination in VLMs has centered on training-stage and inference-stage approaches. Training-stage approaches, such as robust instruction tuning \cite{Jiang_2024_CVPR, DBLP:conf/cvpr/Yu0WPYQT0Z24, 10656676}, enhancing output consistency and reliability but incur substantial training and data costs. Inference-stage approaches include post-hoc techniques with auxiliary networks \cite{DBLP:conf/acl/FengS00BT24, wu-etal-2024-logical, yin2024woodpecker} and diverse decoding strategies \cite{10655750, DBLP:conf/cvpr/HuangDZ0H0L0Y24, Kim2023InstructiveDI, 10657718, wang-etal-2024-mitigating, chuang2024dola} to suppress spurious content. Although effective, post-hoc techniques introduce additional inference overhead, and decoding strategies often exhibit limited generalization.
Recent studies also focus on abnormal behaviors related to object hallucination, such as excessive focus on special tokens \cite{DBLP:conf/cvpr/HuangDZ0H0L0Y24, tang2025seeing, tang2025intervening} or insufficient attention to visual tokens \cite{10657718, DBLP:conf/iclr/YangL0X25}, and attempt to alleviate them through targeted regulation. Our method, \textbf{AIR}, falls under inference-stage approaches, but unlike existing techniques, it avoids introducing additional inference overhead while maintaining strong generalization performance.

\section{Preliminary}

\textbf{Notation.}
Vectors of all zeros and ones are denoted by $\mathbf{0}$ and $\mathbf{1}$, respectively. 
The $i$-th one-hot vector is written as $\mathbf{e}_i$. 
A vector is represented in boldface, e.g., $\mathbf{a}$, with its $i$-th scalar element denoted by $a_i$. 
A matrix is denoted by a capital boldface letter, e.g., $\mathbf{A}$, and $\mathbf{A}_i$ refers to its $i$-th column vector, unless otherwise specified.

\noindent\textbf{Transformer.} 
Let $\mathbf{X} := [\mathbf{x}_1, \mathbf{x}_2, \dots, \mathbf{x}_T] \in \mathbb{R}^{d \times T}$ denote an input sequence of $T$ tokens with $d$ dimensions, 
where each token represents a single element of the input modality, such as a textual or visual token.
The $\ell$-th (single-head) self-attention layer is defined as:
\begin{align*}
	\mathbf{A}^\ell &= S\!\left(\frac{(\mathbf{X}^{\ell-1})^\top \mathbf{W}_{\mathrm{QK}} \mathbf{X}^{\ell-1}}{\sqrt{d}}\right),  \\
	\mathbf{U}^\ell &= \mathbf{W}_V \mathbf{X}^{\ell-1} \mathbf{A}^\ell,
\end{align*}
where $\mathbf{W}_V \in \mathbb{R}^{d \times d}$ is the value matrix, 
$\mathbf{W}_{\mathrm{QK}} := \mathbf{W}_Q^\top \mathbf{W}_K \in \mathbb{R}^{d \times d}$ is the query-key parameter matrix, 
and $S : \mathbb{R}^T \to \mathbb{R}^T$ denotes the softmax function. 
In this way, each input token in $\mathbf{X}^{\ell-1}$ (embedded by $\mathbf{W}_V$) is mixed by $\mathbf{A}^\ell$.
Then the transformer block is defined as:
\begin{align*}
	\mathbf{Z}^\ell &:= \mathbf{U}^\ell + \mathbf{X}^{\ell-1}, \\
	\mathbf{H}^\ell &:= \mathbf{W}_{F_2}\,\sigma(\mathbf{W}_{F_1} \mathbf{Z}^\ell), \\
	\mathbf{X}^\ell &:= \mathbf{H}^\ell + \mathbf{Z}^\ell,
\end{align*}
where $\mathbf{H}^\ell$ denotes a feed-forward network parameterized by 
$\mathbf{W}_{F_1}, \mathbf{W}_{F_2} \in \mathbb{R}^{d \times d}$ 
and an element-wise activation function $\sigma : \mathbb{R} \to \mathbb{R}$.
For simplicity, we omit the token embedding layer and set $\mathbf{X}^0 := \mathbf{X}$. 
Layer normalization is applied after the residual connections (i.e., on $\mathbf{Z}^\ell$ and $\mathbf{X}^{\ell}$) 
and before the inputs, respectively.
Finally, the transformer block is stacked $L$ times, yielding the output
$\mathbf{F}(\mathbf{X}) := \mathbf{X}^L \in \mathbb{R}^{d \times T}$.
In the multi-head setting, we define $\mathbf{A}^{(h,\ell)}$ as the attention matrix associated with the $h$-th head in the $\ell$-th layer, where $\mathbf{A}_{i,j}$ represents the attention weight from the $i$-th token to the $j$-th token.\footnote{Here we omit the supscript when the context is clear.}

\section{Attention Imbalance}\label{4}

\subsection{Formulation of Attention Imbalance} 
Imbalanced attention allocation manifests at different granularity levels in LVLMs, reflecting unequal distribution of attention across modalities and among tokens within a sequence. To characterize this phenomenon, we introduce a concept termed \textit{attention imbalance}, which is composed of two complementary forms: \textit{Modality-Wise Attention Imbalance} (MAI) and \textit{Token-Wise Attention Imbalance} (TAI).

\noindent\textbf{Modality-wise Attention Imbalance (MAI).}  
MAI measures the degree of uneven attention allocation across different modalities in multimodal inputs.

Assume the input consists of a set of modalities, i.e., $\mathcal{M} = \{M_1, M_2, \dots, M_k\}$, where each modality $M_p$ contains $|M_p|$ tokens.  
Let $\mathcal{I}_p$ denote the index set of tokens belonging to modality $M_p$:
\[
M_p = \{\mathbf{x}_j \mid j \in \mathcal{I}_p\}, \quad \bigcup_{p=1}^k \mathcal{I}_p = \{1, \dots, T\}.
\]

Let $\mathbf{A} \in \mathbb{R}^{T \times T}$ denote the attention matrix. The total attention mass received by modality $M_p$ is defined as:
\[
A_{M_p} := \sum_{j \in \mathcal{I}_p} \sum_{i=1}^{T} \mathbf{A}_{i,j}.
\]



For any two modalities $M_p$ and $M_q$, MAI is defined as:
\[
\mathrm{MAI}(M_p, M_q) := \frac{A_{M_p}}{A_{M_q}}.
\]

\noindent\emph{Intuition}. MAI is used to quantify the ratio of the overall attention assigned to the two modalities.

\noindent\textbf{Interpretation of MAI.}
\begin{itemize}
	\item $\mathrm{MAI}(M_p, M_q) \approx 1$: modalities $M_p$ and $M_q$ receive comparable attention;
	\item $\mathrm{MAI}(M_p, M_q) \gg 1$: modality $M_p$ dominates the attention distribution;
	\item $0 < \mathrm{MAI}(M_p, M_q) \ll 1$: modality $M_q$ dominates the attention distribution.
\end{itemize}

\noindent\textbf{Token-wise Attention Imbalance (TAI).} TAI quantifies the mismatch between the model’s attention allocation and the actual information contribution of individual tokens within a sequence.

Let $\mathbf{A} \in \mathbb{R}^{T \times T}$ denote the attention matrix. 
The total attention received by token $\mathbf{x}_j$ is denoted by:
\[
A(\mathbf{x}_j) := \sum_{i=1}^{T} \mathbf{A}_{i,j}.
\]

Given $i$ tokens, we apply the conditional mutual information 
$I(\mathbf{x}_j; \mathbf{x}_{i+1} \mid \mathbf{x}_{\le i})$ to measure how much each preceding token $\mathbf{x}_j$ ($j\le i$) contributes to predicting the next target token \(x_{i+1}\) \cite{tang2025seeing}. Then the TAI of $\mathbf{x}_j$ is defined as:
\[
\mathrm{TAI}(\mathbf{x}_j) := 
\frac{A(\mathbf{x}_j)}{\sum_{k=1}^{i} A(\mathbf{x}_k)}
\;\cdot\;
\frac{\sum_{k=1}^{i} I(\mathbf{x}_k; \mathbf{x}_{i+1} \mid \mathbf{x}_{\le i})}{I(\mathbf{x}_j; \mathbf{x}_{i+1} \mid \mathbf{x}_{\le i})}.
\]

\noindent\emph{Intuition}. TAI is used to quantify the ratio between the attention proportion allocated to a token and its proportion of information contribution.

\noindent\textbf{Interpretation of TAI.}  
\begin{itemize}
	\item $\mathrm{TAI}(\mathbf{x}_j) \approx 1$: the attention allocated to token $\mathbf{x}_j$ is approximately proportional to its information contribution.  
	\item $\mathrm{TAI}(\mathbf{x}_j) \gg 1$: token $\mathbf{x}_j$ is over-attended relative to its information contribution.
	\item $0 < \mathrm{TAI}(\mathbf{x}_j) \ll 1$: token $\mathbf{x}_j$ is under-attended relative to its information contribution.  
\end{itemize}

\begin{figure}[t]
	\centering
	\includegraphics[width=0.475\textwidth]{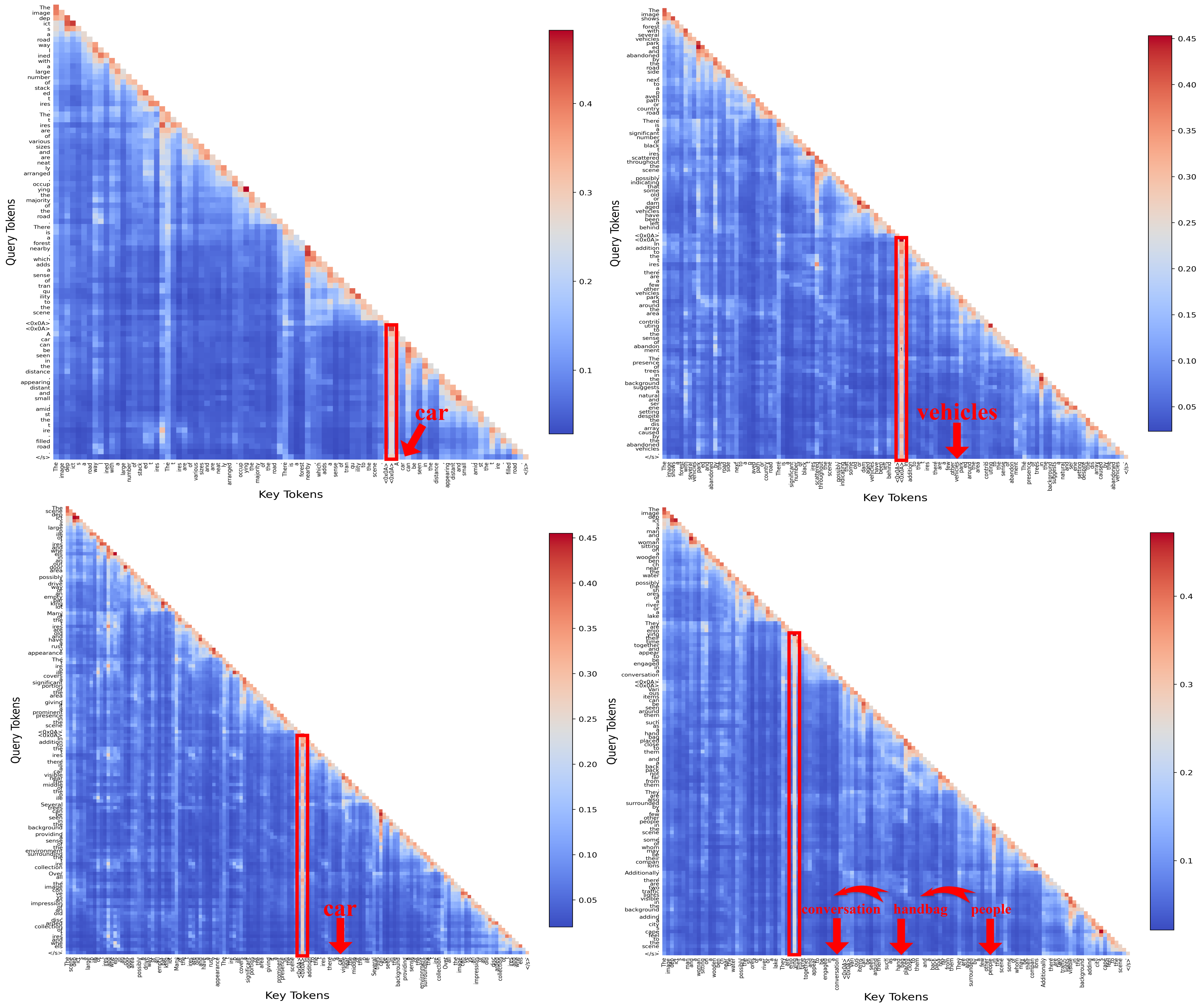}
	\caption{The attention map of output tokens. Imbalanced token is highlighted with a red box, and the hallucinated tokens are indicated by red arrows \textcolor{red}{$\downarrow$}.}
	\label{fig1}
\end{figure}

\subsection{Attention Imbalance in Object Hallucination}
We systematically investigate the underlying attention imbalance of object hallucination. In Section \ref{4.2.1}, we demonstrate a significant co-occurrence between TAI and object hallucination. In Section \ref{4.2.2}, we move to the attention-head level, finding that hallucination-sensitive heads exhibit more pronounced MAI. Subsequently, we reveal a potential underlying cause of this disparity: hallucination-sensitive heads tend to inherit closer attention patterns from the base language model. We choose the widely adopted LLaVA-1.5-7B \cite{10655294} for demonstration.

\subsubsection{Co-occurrence between TAI and Object Hallucination}\label{4.2.1}
We primarily focus on over-attended imbalanced tokens with high TAI that easily drive object hallucination.
To identify such tokens, we set a threshold \( \tau \) and select those whose TAI values exceed $\tau$. 

\noindent\textbf{Computation of threshold $\tau$}. For each example in the evaluation dataset, we compute the TAI values of all generated tokens and take the maximum as its representative value. We then collect these maximum TAI values across all examples, compute their mean \( \mu \) and standard deviation \( \sigma \), and define the threshold as $\tau = \mu + \sigma$. 

\begin{figure}[t]
	\centering
	\includegraphics[width=0.42\textwidth]{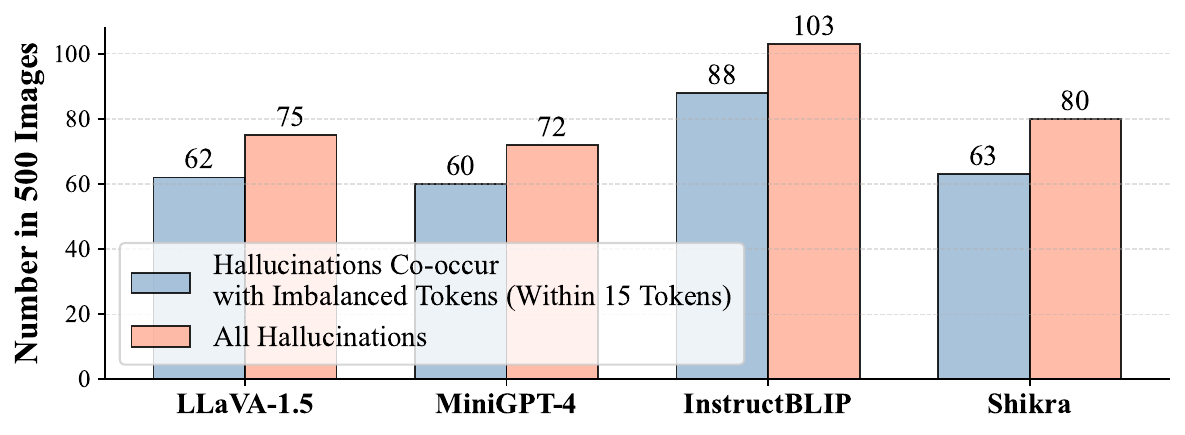}
	\caption{Co-occurrence between imbalanced tokens and object hallucinations across LVLMs.}
	\label{fig2}
\end{figure}

\noindent\textbf{Co-occurrence}. Experimental results reveal a significant co-occurrence between the emergence of such tokens and object hallucination: prior to hallucination generation, the model typically produces 
at least one imbalanced token, and then generates the hallucinated content within the following $15$ tokens.
Figure~\ref{fig1} shows that hallucinated words such as \textit{car} and \textit{vehicles} appear immediately after the imbalanced token \textit{\textless 0x0A\textgreater} (with a markedly TAI value of 98 surpassing the threshold $\tau=21$). In addition, we observe a hallucination propagation phenomenon where words like \textit{handbag} and \textit{people} incur \textit{conversation}, which aligns with the snowball-type hallucination in recent work~\cite{tang2025seeing}.
Figure \ref{fig2} illustrates that this co-occurrence is consistently observed across four representative LVLMs: LLaVA-1.5, MiniGPT-4, InstructBLIP, and Shikra. 
We argue that the model’s excessive focus on high-TAI tokens serves as a key factor contributing to object hallucination.

\begin{figure*}[t]
	\centering
	\includegraphics[width=0.95\textwidth]{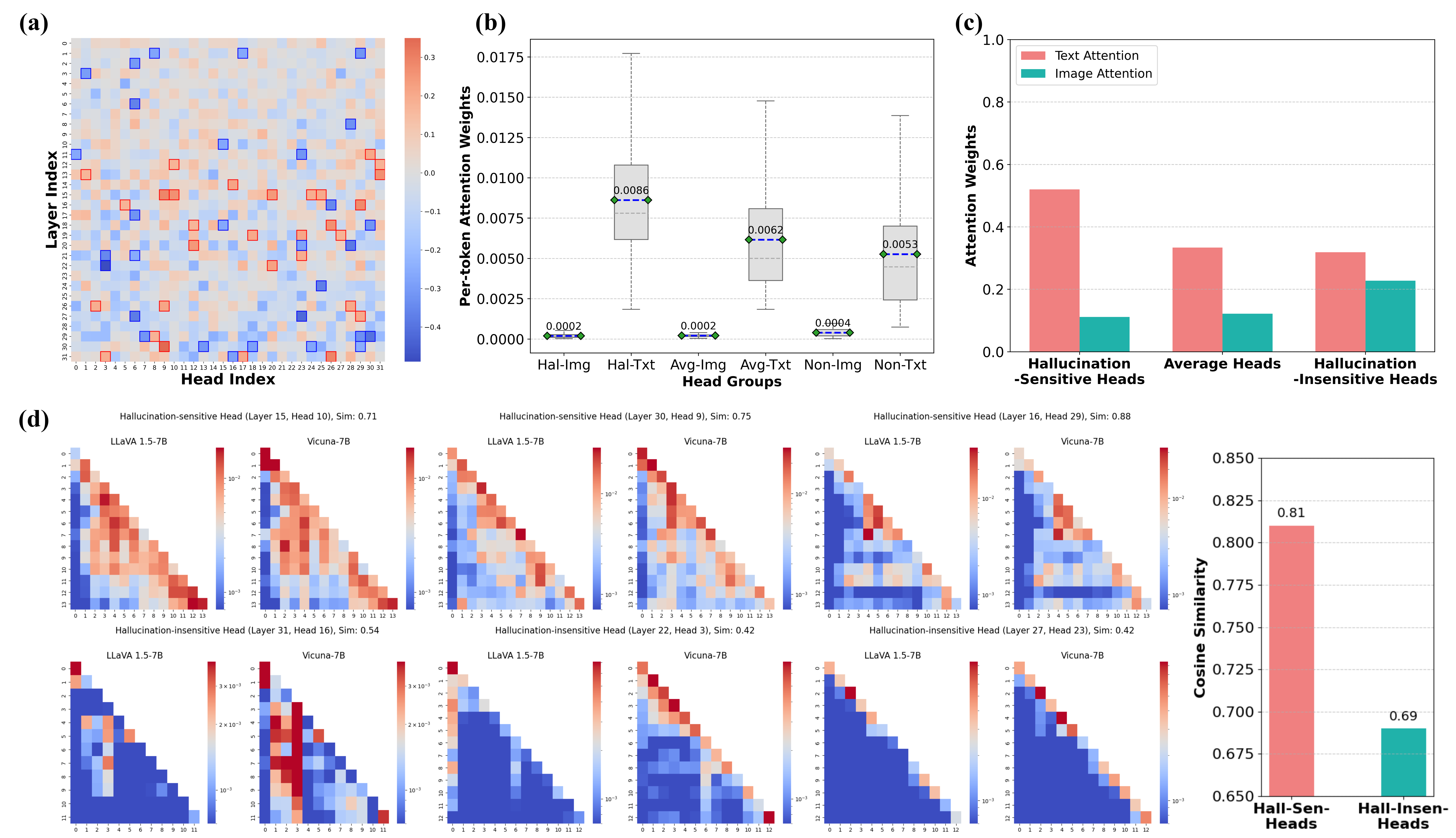}
	\caption{(a) The distribution of attention heads. We highlight the top twenty hallucination-sensitive heads with red boxes and the bottom twenty hallucination-insensitive heads with blue boxes. (b) Comparison of the average modality-token attention weights among hallucination-sensitive heads, average heads, and hallucination-insensitive heads. (c) Comparison of the total modality-level attention weights of three head groups. (d) Comparison of attention map similarities between the LVLM and its base language model. }
	\label{fig3}
    \vspace{-3mm}
\end{figure*}

\subsubsection{High MAI in Hallucination-Sensitive Attention Heads}\label{4.2.2}
By the erasure-based attribution method \cite{voita-etal-2019-analyzing, wiegreffe-pinter-2019-attention, clark-etal-2019-bert, abnar-zuidema-2020-quantifying}, a technique that evaluates the importance of model components by removing them and observing changes in predictions, we erase individual attention heads to measure their impact on the probability of generating hallucinated tokens (more details are provided in Appendix~A.1).
According to the change of the hallucination probabilities before and after erasure, we categorize attention heads into three types:

\begin{itemize}
    \item \textbf{Hallucination-sensitive heads:}  
    Erasing these heads leads to a significant reduction in the probability of generating hallucinated tokens, indicating their strong influence on hallucination production.

    \item \textbf{Hallucination-insensitive heads:}  
    Erasing these heads has little or no effect on the probability of hallucinated tokens, suggesting that they play a limited role in hallucination generation.

    \item \textbf{Average heads:}  
    The average effect computed over all attention heads, serving as a reference baseline.
\end{itemize}

Figure~\ref{fig3} (a) reveals that \textit{hallucination-sensitive heads are primarily concentrated in the middle and upper layers of the model}, with the highest density observed in the middle layers. This observation is consistent with previous research that highlights the middle layers' crucial role in hallucination generation \cite{jiang2024devils}.

\noindent\textbf{Attention imbalance across head types.}
We compare the three types of attention heads (each consisting of twenty heads): hallucination-sensitive heads, hallucination-insensitive heads, and average heads, to analyze attention imbalance. 
Figure~\ref{fig3} (b) shows the per-token average attention distribution between textual and visual tokens. Across all $60$ heads, textual tokens consistently receive higher average attention than visual tokens, with the disparity being most pronounced in hallucination-sensitive heads. 
We further examine the total cross-modal attention imbalance.

As illustrated in Figure~\ref{fig3} (c), the overall attention imbalance gradually decreases from hallucination-sensitive heads to average heads and further to hallucination-insensitive heads. Hallucination-sensitive heads exhibit an MAI of 5.1, substantially higher than that of hallucination-insensitive heads (i.e., 1.5), while average heads reach an intermediate MAI of 2.4. These findings suggest that hallucination-sensitive heads allocate a disproportionately larger share of attention to textual information, indicating a stronger text-to-image bias, whereas hallucination-insensitive heads show comparatively greater attention toward visual information.

\noindent\textbf{Hallucination-sensitive heads inherit stronger textual attention patterns from the base language model.}
We observe that hallucination-sensitive heads display a more pronounced attention imbalance. Specifically, these heads tend to over-emphasize the textual modality and disproportionately allocate attention to high-MAI tokens within the sequence.
We propose a potential hypothesis that these heads may inherit stronger attention patterns from the base language model compared to other groups.
Specifically, we compare LLaVA-7B with its base language model Vicuna-7B. 
As shown in Figure~\ref{fig3} (d), the cosine similarity analysis reveals a strong alignment between the attention maps of tokens produced by the hallucination-sensitive heads in LLaVA-7B and those in Vicuna-7B (cosine score = 0.81). In contrast, the hallucination-insensitive heads show a weaker correspondence (cosine score = 0.69). These results provide empirical support for our hypothesis. Additional computational details are provided in Appendix A.2.

\section{Mitigating Object Hallucination} \label{5}
We propose \textbf{A}ttention \textbf{I}mbalance \textbf{R}ectification (\textbf{AIR}), a method designed to mitigate hallucinations by rectifying attention imbalance. 
In Section~\ref{sec5.1}, we introduce \textit{modality-balanced attention reallocation}, which leverages text attention from hallucination-sensitive heads as a signal to reallocate cross-modal attention. 
This mechanism guides the model to reduce its overreliance on textual information and to strengthen its focus on visual semantics. 
Section~\ref{sec5.2} then presents \textit{variance-constrained projection regularization}, which adjusts the variance of the attention matrix to balance attention distributions, thereby suppressing the model’s tendency to over-emphasize attention-imbalanced tokens. 

\begin{algorithm}[t]
	\small
	\caption{AIR: Attention Imbalance Rectification}
	\label{air}
	\begin{algorithmic}[1]
		\Require Hallucination-sensitive head set $\mathcal{H}_{\mathrm{sens}}$, 
		threshold $\tau_{\mathrm{text}}$, reallocation factors $\lambda$, $\gamma$, and regularization coefficient $\xi$, $\beta$.
		\State $\mathbf{W}_{\mathrm{QK}} \leftarrow (1 - \xi\frac{1}{\log(\mathrm{tr}(\mathbf{W}_{\mathrm{QK}}^2) + 10^{-6})})\mathbf{W}_{\mathrm{QK}}$
		\For{each decoding step $t$}
		\For{head $(\ell,h) \in \mathcal{H}_{\mathrm{sens}}$}
		\State // \textit{Modality-balanced attention reallocation}
		\State Compute text-attention value $V^{\mathrm{text}}_{(\ell,h)}$
		\If {$V^{\mathrm{text}}_{(\ell,h)} > \tau_{\mathrm{text}}$}
		\State 
		\[
		A^{\prime(\ell,h)}_{ij} \leftarrow 
		\begin{cases}
			{\lambda}A^{(\ell,h)}_{ij}, & j \in \mathcal{T},\\[2pt]
			{\gamma}A^{(\ell,h)}_{ij}, & j \in \mathcal{V},
		\end{cases}
		\]
		\Else
		\State $A^{\prime(\ell,h)} \leftarrow A^{(\ell,h)}$
		\EndIf
		\State // \textit{Variance-constrained projection regularization}
		\State 
		$\hat{A}^{(\ell,h)} \leftarrow 
		A^{\prime(\ell,h)} - \frac{\mathrm{tr}(A^{\prime(\ell,h)})}{L}\mathbf{I}$
		\State 
		$\tilde{A}^{(\ell,h)} \leftarrow 
		\hat{A}^{(\ell,h)} \cdot 
		\sqrt{\frac{\|A^{\prime(\ell,h)}\|_F^2}
			{\|\hat{A}^{(\ell,h)}\|_F^2 + \varepsilon}}$
		\State \[
		A^{\ast(\ell,h)} \leftarrow (1-\beta)\tilde{A}^{(\ell,h)} 
		+ \beta\,\mathrm{mean}(\tilde{A}^{(\ell,h)})\,\mathbf{1}
		\]
		\State Perform self-attention update.
		\EndFor
		\EndFor
	\end{algorithmic}
\end{algorithm}

\subsection{Modality-Balanced Attention Reallocation}\label{sec5.1}
We leverage the over-attention signals from hallucination-sensitive heads during decoding to regulate cross-modal attention. The goal is to correct excessive modality-wise imbalance, reducing the model’s reliance on the textual modality while enhancing its focus on visual information. Attention reallocation is detailed in Lines 2–10 of Algorithm \ref{air}.

At each decoding step \(t\), for every hallucination-sensitive attention head 
\[
\mathrm{head} = (\ell, h), \quad \mathrm{head} \in \mathcal{H}_{\mathrm{sens}},
\]
where \(\ell\) denotes the layer index and \(h\) the head index within that layer, 
the causal self-attention matrix is computed as:
\[
\mathbf{A}^{(\ell,h)} = S\!\left(\frac{(\mathbf{X}^{\ell-1})^{\top} \mathbf{W}_{\mathrm{QK}}^{(\ell,h)} \mathbf{X}^{\ell-1}}{\sqrt{d}}\right),
\]
where \(S(\cdot)\) denotes the softmax function and 
\(\mathbf{W}_{\mathrm{QK}}^{(\ell,h)}\) is the query–key parameter matrix of the \(h\)-th head in layer \(\ell\).
The cumulative attention value allocated to textual tokens is defined as:
\[
V^{\mathrm{text}}_{(\ell,h)} = \sum_{i \in \mathcal{T}} \sum_{j \in \mathcal{T}} A^{(\ell,h)}_{ij},
\]
where \(\mathcal{T}\) denotes the index set of textual tokens. 
When \(V^{\mathrm{text}}_{(\ell,h)} > \tau_{\mathrm{text}}\) (with \(\tau_{\mathrm{text}}\) denoting the imbalance threshold), 
the head is considered over-attentive to text.

We then apply a cross-modal reallocation defined by:
\[
A^{\prime (\ell,h)}_{ij} =
\begin{cases}
	\lambda\,A^{(\ell,h)}_{ij}, & j \in \mathcal{T},\\[4pt]
	\gamma\,A^{(\ell,h)}_{ij}, & j \in \mathcal{V},
\end{cases}
\]
where \(\lambda \in [0,1]\) controls the degree of text suppression, 
\(\gamma > 1\) is a hyperparameter that amplifies visual attention, 
and \(\mathcal{V}\) denotes the set of visual tokens. 
This mechanism reallocates attention within hallucination-sensitive heads to achieve coordinated text-suppression and visual-amplification control across layers, effectively mitigating modality-wise attention imbalance.

\subsection{Variance-Constrained Projection Regularization}\label{sec5.2}
To alleviate excessive attention concentration and enhance the stability of attention behavior during decoding, we first apply a lightweight adjustment on the projection weights $\mathbf{W}_{\mathrm{QK}}$, 
which adaptively rescales the query-key mapping according to its spectral energy (line 1 of Algorithm \ref{air}).

Building upon this, we introduce a projection regularization that explicitly controls the variance of the attention matrix $\mathbf{A}^{(\ell,h)}$. 
By constraining the variability of $\mathbf{A}^{(\ell,h)}$, the model learns to produce a more balanced and uniform attention distribution, thereby reducing token-wise imbalance.
A detailed mathematical analysis of the relationship between $\mathbf{A}^{(\ell,h)}$, $\mathbf{W}_{\mathrm{QK}}$ and the resulting attention distribution is provided in Appendix B.
Projection regularization is detailed in Lines 12–14 of Algorithm \ref{air}.

For the self-attention matrix $\mathbf{A}^{(\ell,h)}$, a zero-trace projection is applied to remove global diagonal bias:
\[
\hat{\mathbf{A}}^{(\ell,h)} = \mathbf{A}^{(\ell,h)} - 
\frac{\mathrm{tr}(\mathbf{A}^{(\ell,h)})}{L}\,\mathbf{I},
\]
where $L$ denotes the sequence length and $\mathbf{I}$ the identity matrix.  
This operation ensures that the mean self-attention strength across tokens is neutralized,  
thereby discouraging the model from over-relying on self-aligned activations.  

To prevent the projected scores from collapsing or exploding,  
we further rescale them to preserve Frobenius energy:
\[
\tilde{\mathbf{A}}^{(\ell,h)} = \hat{\mathbf{A}}^{(\ell,h)} \cdot 
\sqrt{\frac{\|\mathbf{A}^{(\ell,h)}\|_F^2}
	{\|\hat{\mathbf{A}}^{(\ell,h)}\|_F^2 + \varepsilon}},
\]
where $\varepsilon$ is a small constant (typically $10^{-8}$) to avoid numerical instability.  
Finally, a shrinkage regularization is introduced to reduce excessive dispersion while maintaining representational diversity:
\[
\mathbf{A}^{\ast(\ell,h)} = (1-\beta)\,\tilde{\mathbf{A}}^{(\ell,h)} + 
\beta\,\mathrm{mean}(\tilde{\mathbf{A}}^{(\ell,h)})\,\mathbf{1},
\]
where $\beta \in [0,1]$ controls the degree of shrinkage toward the mean matrix $\mathbf{1}$.  
The resulting $\mathbf{A}^{\ast(\ell,h)}$ is then used for the subsequent attention update computation.

\begin{table*}[t]
	\centering
	\caption{\textbf{CHAIR} hallucination evaluation results on four LVLMs across baselines. Smaller \(C_S\) and \(C_I\) indicate fewer hallucinations. Results are reported with maximum new tokens set to 256 and 64. Our method AIR achieves the best hallucination mitigation performance across different max new token settings compared to the baselines. $\Delta\%$ represents the relative improvement in performance over the second-best method.}
	\label{tab:chair_results}
	\setlength{\tabcolsep}{2.3pt}
	\footnotesize
	\begin{threeparttable}
		\begin{tabular}{lcccccccc|cccccccc}
			\toprule
			\rowcolor{gray!10} & \multicolumn{8}{c|}{\textbf{Max New Tokens: 256}} &
			\multicolumn{8}{c}{\textbf{Max New Tokens: 64}} \\
			\cmidrule(lr){1-9}\cmidrule(lr){10-17}
			\multirow{2}{*}{\textbf{Methods}}
			& \multicolumn{2}{c}{LLaVA-1.5} & \multicolumn{2}{c}{MiniGPT4}
			& \multicolumn{2}{c}{InstructBLIP}     & \multicolumn{2}{c|}{Shikra}
			& \multicolumn{2}{c}{LLaVA-1.5} & \multicolumn{2}{c}{MiniGPT4}
			& \multicolumn{2}{c}{InstructBLIP}     & \multicolumn{2}{c}{Shikra} \\
			& \(C_S\!\downarrow\) & \(C_I\!\downarrow\) & \(C_S\!\downarrow\) & \(C_I\!\downarrow\)
			& \(C_S\!\downarrow\) & \(C_I\!\downarrow\) & \(C_S\!\downarrow\) & \(C_I\!\downarrow\)
			& \(C_S\!\downarrow\) & \(C_I\!\downarrow\) & \(C_S\!\downarrow\) & \(C_I\!\downarrow\)
			& \(C_S\!\downarrow\) & \(C_I\!\downarrow\) & \(C_S\!\downarrow\) & \(C_I\!\downarrow\)
			\\
			\midrule
			Greedy   & 51.8 & 13.7 & 43.0 & 13.4 & 54.7 & 20.2 & 55.4 & 15.2
			& 20.8 & 6.2 & 28.8 & 11.4 &29.8 & 14.2 & 22.3  &8.2   \\
			FarSight & 43.9  & 12.5  & 39.2  & 12.4  & 48.3  & 18.7  & 51.6  & 12.2 
			& 18.6  & 5.9  &  22.5 & 8.9  & 20.2  & 9.6  & 15.4  & 7.6  \\
			VCD   & 59.4 & 16.0 & 40.0 & 14.6 & 50.1 & 19.4 & 54.9 & 15.1
			& 22.8 & 7.1  & 28.0 & 11.3 & 34.1 & 15.8 & 23.5 & 8.3  \\
			DoLA    & 54.0 & 14.2 & 42.0 & 13.5 & 52.8 & 19.1 & 55.7 & 14.8
			& 22.2 & 6.7  & 28.6 & 11.5 & 28.1 & 14.2  & 20.2 & 9.6  \\
			HALC     & 51.4 & 13.1 & 37.4 & 12.1 & 48.0 & 15.3 & 47.2 & 14.5
			& 21.2 & 6.6 &  24.2 & 9.7  & 24.2  & 14.1  & 14.4  &7.8   \\
			OPERA    & 44.1 & 12.8 & 37.3 & 13.5 & 48.2 & 14.3 & 38.2 & 14.2
			& 17.8 & 6.5 &  26.6 & 10.4  & 18.4  & 8.7  & 14.6  & 7.4  \\
			AD-HH      & 35.2 & 8.8 & 32.8 & 11.5 & 36.0 & 10.3 & 36.9 & 13.7
			& 15.6 & 5.7  & 22.0 & 8.5 & 19.6 & 8.5  & 13.8 & 7.0  \\
			\rowcolor{gray!10}
			\textbf{AIR (Ours)} & \textbf{28.8} & \textbf{8.6} & \textbf{21.3} & \textbf{8.9} & \textbf{30.1} & \textbf{8.6} & \textbf{30.3} & \textbf{9.3}
			& \textbf{14.8} & \textbf{5.1} & \textbf{17.2}  & \textbf{7.5}  & \textbf{16.3}  & \textbf{8.1}  & \textbf{12.9}  &\textbf{7.0}   \\
			
			$\Delta\%$
			& \textcolor{Blue}{$\downarrow$18.1\%} & \textcolor{Blue}{$\downarrow$2.3\%} & \textcolor{Blue}{$\downarrow$35.1\%} & \textcolor{Blue}{$\downarrow$22.6\%}
			& \textcolor{Blue}{$\downarrow$16.4\%} & \textcolor{Blue}{$\downarrow$16.5\%} & \textcolor{Blue}{$\downarrow$17.9\%} & \textcolor{Blue}{$\downarrow$23.8\%}
			& \textcolor{Blue}{$\downarrow$5.1\%} & \textcolor{Blue}{$\downarrow$10.5\%} & \textcolor{Blue}{$\downarrow$21.8\%} & \textcolor{Blue}{$\downarrow$11.8\%}
			& \textcolor{Blue}{$\downarrow$16.8\%} & \textcolor{Blue}{$\downarrow$4.7\%} & \textcolor{Blue}{$\downarrow$6.5\%} & \textcolor{Blue}{$\downarrow$5.4\%} \\
			
			\bottomrule
		\end{tabular}
	\end{threeparttable}
\end{table*}

\begin{figure*}[t]
	\centering
	\includegraphics[width=0.98\textwidth]{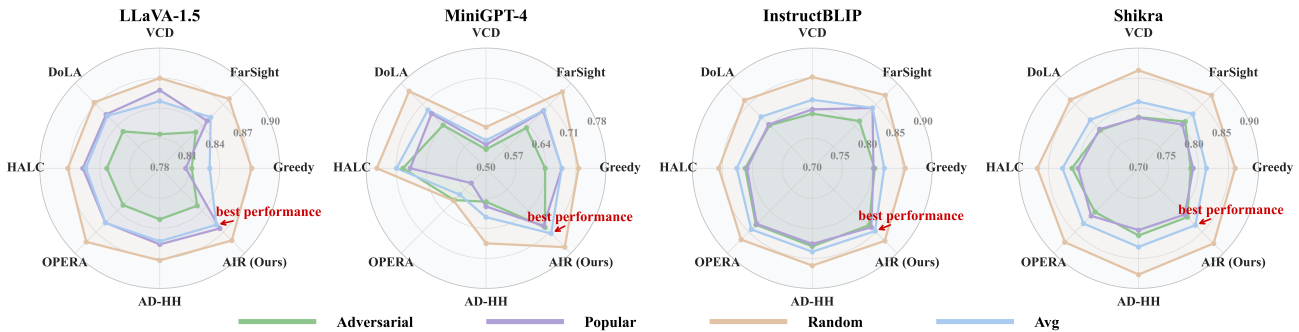}
	\caption{\textbf{POPE} hallucination evaluation results of four LVLMs across different baselines. The four filled regions denote Adversarial, Popular, Random subsets and their average. AIR consistently achieves the highest average performance across all four LVLMs.}
	\label{fig4}
\end{figure*}

\section{Experiments}
\subsection{Experimental Setup}
\textbf{Models and Baselines.}
We evaluate four of the most representative LVLMs, including LLaVA-1.5 \cite{10655294}, MiniGPT-4 \cite{zhu2024minigpt}, InstructBLIP \cite{dai2023instructblipgeneralpurposevisionlanguagemodels}, and Shikra~\cite{Chen2023ShikraUM}.
Since our work focuses on decoding-time approaches for LVLMs, we adopt standard greedy decoding as the naive baseline and further include six state-of-the-art decoding-time methods for comparison: FarSight \cite{tang2025seeing}, VCD \cite{10657718}, DoLA \cite{chuang2024dola}, HALC~\cite{pmlr-v235-chen24bi}, OPERA \cite{DBLP:conf/cvpr/HuangDZ0H0L0Y24}, and AD-HH \cite{DBLP:conf/iclr/YangL0X25}. Additional experimental analyzes are presented in Appendix C.


\noindent\textbf{Datasets and  Metrics.}
We evaluate our approach on both hallucination and general LVLM benchmarks. For hallucination evaluation, we employ CHAIR~\cite{rohrbach-etal-2018-object} and POPE \cite{li2023evaluating}, both derived from the MSCOCO dataset \cite{10.1007/978-3-319-10602-1_48}.
For CHAIR, each LVLM is prompted with ``\textit{Please describe this image in detail.}''. The benchmark quantifies the proportion of hallucinated objects that appear in model-generated captions but are absent from ground-truth annotations. It includes two hallucination metrics: sentence-level (\(\mathrm{CHAIR_S}\)) and image-level (\(\mathrm{CHAIR_I}\)) hallucination rates where lower values indicate fewer hallucinations:
\[
\begin{aligned}
	C_S &= 
	\frac{|\{\text{captions with hallucinated objects}\}|}
	{|\{\text{all captions}\}|},\\[3pt]
	C_I &=
	\frac{|\{\text{hallucinated objects}\}|}
	{|\{\text{all mentioned objects}\}|}.
\end{aligned}
\]

The POPE benchmark further evaluates object-level hallucination through a question--answering format. Each LVLM is queried with prompts such as ``\textit{Is there a \textless object\textgreater\ in the image?}''
to test its reliability in object recognition. POPE consists of three evaluation splits: Random, Popular, and Adversarial, corresponding to randomly sampled objects, frequently occurring objects, and visually or semantically confusable objects, respectively.
For general vision--language capability, we employ the MM-Vet benchmark \cite{yu2024mm}, which assesses the overall performance of MLLMs across six open-ended reasoning domains: recognition (Rec), language generation (Gen), optical character recognition (OCR), spatial reasoning (Spat), knowledge understanding (Know), and mathematical reasoning (Math).

\noindent\textbf{Implementation Details.}  
Our AIR method selects the top 20 hallucination-sensitive attention heads for each model to form \(\mathcal{H}_{\mathrm{sens}}\).  
The text-attention threshold \(\tau_{\mathrm{text}}\), modulation factors \(\lambda\) and \(\gamma\), regularization coefficient \(\xi\), and \(\beta\) are set to \(0.3\), \(0.1\), \(3.5\), \(0.01\), and \(0.3\), respectively.
The default sequence length of LVLMs is set to 256.
The details of other baselines are provided in Appendix C.1.

\subsection{Experimental Results}
\textbf{AIR significantly reduces object hallucinations up to 35.1\% by rectifying the attention imbalance.} Table~\ref{tab:chair_results} shows that AIR consistently outperforms all baselines across the four LVLMs in the CHAIR evaluation. 
In particular, for LLaVA-1.5 and MiniGPT-4, our method achieves substantial improvements, 
reducing the \(C_S\) score to 28.8 and 21.3, and the \(C_I\) score to 8.6 and 8.9, respectively.
We observe that AIR consistently alleviates hallucinations under different maximum token generation limits (256 and 64), achieving the best hallucination mitigation performance among all baselines and demonstrating strong robustness. Specifically, in MiniGPT4, when the \textit{Max New Tokens} is 256, AIR attains \(C_S\) of 21.3 and \(C_I\) of 8.9, yielding up to 35.1\% and 22.6\% improvements than the second-best baseline; when the limit is 64, it still maintains strong performance with \(C_S\) of 17.2 and \(C_I\) of 7.5.

Figure \ref{fig4} illustrates the performance of AIR on the POPE evaluation. We observe that, except for the Random split of LLaVA-1.5 and the Popular split of Shikra, where AIR performs slightly worse than OPERA, it achieves the best results across the remaining ten splits of the four models. Moreover, AIR attains the highest average scores across all splits (as indicated by the blue line in Figure~\ref{fig4}), demonstrating its effectiveness in mitigating object hallucinations.

\noindent\textbf{The rectification of attention imbalance improves the general capabilities of LVLMs up to 15.9\%.}
Table~\ref{tab:mmvet_llava15} and Table~\ref{tab:mmvet_minigpt4} present the performance of AIR and baseline methods on the general benchmark MM-Vet. A noticeable decline in general capability can be observed across several methods. For instance, in LLaVA-1.5, HALC and AD-HH show decreases of 1.4\% and 14.8\% in overall scores, respectively. This trend is more pronounced in MiniGPT-4, where all methods except AIR and FarSight exhibit varying degrees of performance degradation. In contrast, AIR not only effectively mitigates hallucinations but also maintains and even slightly enhances the model’s general capability (+15.9\% on LLaVA-1.5, +10.0\% on MiniGPT-4).

\noindent\textbf{The essential of attention reallocation and projection regularization to hallucination mitigation.}
As shown in Table~\ref{tab:air_two_module_ablation}, both mechanisms contribute to mitigating hallucinations, and their combination achieves the best overall performance.
Applying only attention reallocation (\textit{R-only}) substantially decreases hallucination rates, with \(C_S\) reduced from 51.8 to 32.1 and \(C_I\) from 13.7 to 9.9, indicating that rebalancing cross-modal attention effectively alleviates text-dominant bias.
The projection regularization mechanism (\textit{P-only}) improves performance by suppressing attention variance and promoting a more uniform attention distribution.  
When both mechanisms are enabled (\textit{Full}), the model achieves \(C_S{=}\!28.8\) and \(C_I{=}\!8.6\) while maintaining strong general capability.  
The sensitivity analysis of AIR’s hyperparameters is provided in Appendix~C.2.

\begin{table}[t]
	\centering
	\caption{\textbf{MM-Vet} evaluation results on \textbf{LLaVA-1.5}. The best scores are highlighted in bold. AIR demonstrates stronger general capability compared with the baseline methods.}
	\label{tab:mmvet_llava15}
	\setlength{\tabcolsep}{2.8pt}
	\renewcommand{\arraystretch}{1}
	\footnotesize
	\begin{tabular}{lccccccc}
		\toprule
		\textbf{Method} & \textbf{Rec} & \textbf{OCR} & \textbf{Know} & \textbf{Gen} & \textbf{Spat} & \textbf{Math} & \textbf{Overall} $\uparrow$ \\
		\midrule
		Greedy  & 32.7 & 17.6 & 14.5 & 15.3 & 21.9 & 7.7 & 27.6 \\
		FarSight & 32.3 & 20.6 & 14.4 & 16.6 & 23.5 & 11.5 & 28.3 (+2.5\%)
		\\
		VCD     & 32.5 & 21.5 & 15.0 & 17.6 & 24.8 & 3.8 & 28.3 (+2.5\%)\\
		DoLA    & 32.1 & 18.3 & 13.8 & 14.6 & 22.0 & 7.7 & 27.6 (+0.0\%)\\
		HALC    & 32.4 & 17.7 & 14.2 & 15.5 & 19.9 & 7.7 & 27.2 (-1.4\%)\\
		OPERA   & 32.9 & 18.4 & 14.4 & 15.5 & 22.3 & 7.7 & 27.9 (+1.1\%)\\
		AD-HH   & 29.3 & 16.8 & 8.2  & 16.5 & \textbf{25.6} & 6.9 & 23.5 (-14.8\%)\\
		\rowcolor{gray!10}
		\textbf{AIR (Ours)}&\textbf{37.1}	&\textbf{22.8}&	\textbf{18.6}&	\textbf{20.5}& 24.9&	\textbf{11.5}&\textbf{32.0 (+15.9\%)} \\
		\bottomrule
	\end{tabular}
\end{table}

\begin{table}[t]
	\centering
	\caption{\textbf{MM-Vet} evaluation results on \textbf{MiniGPT-4}. The best scores are highlighted in bold. AIR preserves the model’s general capability without degradation.}
	\label{tab:mmvet_minigpt4}
	\setlength{\tabcolsep}{2.8pt}
	\renewcommand{\arraystretch}{1}
	\footnotesize
	\begin{tabular}{lccccccc}
		\toprule
		\textbf{Method} & \textbf{Rec} & \textbf{OCR} & \textbf{Know} & \textbf{Gen} & \textbf{Spat} & \textbf{Math} & \textbf{Overall} $\uparrow$\\
		\midrule
		Greedy  & 22.3 & 14.7 & 13.6 & 10.0 & 21.7 & 7.7 & 20.0 \\
		FarSight & 26.3 & 12.3 & 14.5 & 14.6 & 14.7 & 3.5 & 20.3 (+1.5\%)
		\\
		VCD     & 17.1 & 10.0 & 7.9 & 7.8 & 11.7 & 7.3 & 14.2 (-29.0\%)\\
		DoLA    & 20.7 & 14.4 & 13.6 & 10.0 & 21.7 & 7.3 & 18.8 (-6.0\%)\\
		HALC    & 19.4 & 11.9 & 11.3 & 8.6 & 16.5 & 6.5 & 16.6 (-17.0\%)\\
		OPERA   & 20.9 & \textbf{15.1} & 10.5 & 10.5 & 19.5 & \textbf{12.7} & 18.5 (-7.5\%)\\
		AD-HH   & 23.7 & 12.4 & 11.9 & 12.6 & 14.8 & 3.5 & 18.9 (-5.5\%)\\
		\rowcolor{gray!10}
		\textbf{AIR (Ours)} &25.1&15.1	&\textbf{15.5}&11.2	&\textbf{23.7}&7.7&\textbf{22.0 (+10.0\%)} \\
		\bottomrule
	\end{tabular}
\end{table}

\begin{table}[t]
	\centering
	\caption{Effect of the two core mechanisms in AIR on LLaVA-1.5. \textit{R-only} applies only the attention reallocation, while \textit{P-only} applies only the projection regularization. \textit{Full} enables both mechanisms.}
	\label{tab:air_two_module_ablation}
	\setlength{\tabcolsep}{6pt}
	\footnotesize
	\begin{tabular}{lcccc}
		\toprule
		\textbf{Method} &  \(C_S\!\downarrow\) &  \(C_I\!\downarrow\) & \(F_1\!\uparrow\) & \textbf{MM-Vet Overall}\,$\uparrow$\\
		\midrule
		None  &51.8  &13.7  &0.83  &27.6   \\
		\textit{R-only}     &   32.1  & 9.9 & 0.85   & 30.5 (+10.5\%)   \\
		\textit{P-only}     &    48.6 & 12.1 & 0.84   & 28.0 (+1.4\%)  \\
		\rowcolor{gray!10}
		\textit{Full (AIR)}      &\textbf{28.8}  &\textbf{8.6}  &\textbf{0.86}  &\textbf{32.0} (\textbf{+15.9\%})  \\
		\bottomrule
	\end{tabular}
\end{table}

\section{Conclusion}
We have introduced a novel notion of \textit{attention imbalance} and revealed  its correlation with object hallucination in LVLMs.
We proposed a decoding-time method called \textbf{A}ttention \textbf{I}mbalance \textbf{R}ectification (\textbf{AIR}), to alleviate excessive modality imbalance and reduces the interference of imbalanced tokens, leading to hallucination-free outputs.
Extensive experiments showed that our method outperformed existing SOTA methods in mitigating hallucinations while improving LVLMs' general capability.

{
    \small
    \bibliographystyle{ieeenat_fullname}
    \bibliography{main}
}

\clearpage
\onecolumn
\appendix

\setcounter{page}{1}

\section{Attention Imbalance in Object Hallucination}
We provide a detailed explanation of how we identify specialized attention heads (e.g., hallucination-sensitive and hallucination-insensitive heads) in Section~\ref{A.1}, and describe the computational details for quantifying attention-pattern similarity in Section~\ref{A.2}.

\subsection{Identifying Hallucination-Sensitive and -Insensitive Attention Heads} \label{A.1}
Building on erasure-based attribution \cite{voita-etal-2019-analyzing, wiegreffe-pinter-2019-attention, clark-etal-2019-bert, abnar-zuidema-2020-quantifying}, 
a technique that evaluates the importance of model components by removing them and observing prediction changes, 
We propose a \textit{hallucination-sensitive effect size} to identify attention heads that are either closely linked to hallucinations or largely unrelated.
The goal is to quantify how each head differentially influences hallucinated versus non-hallucinated tokens. 
Formally, let
\[
\mathcal{H} = \{t \mid y_t \in \mathcal{V}_{\mathrm{hall}}\}, \quad 
\mathcal{N} = \{t \mid y_t \in \mathcal{V}_{\mathrm{non}}\},
\]
where $\mathcal{V}_{\mathrm{hall}}$ and $\mathcal{V}_{\mathrm{non}}$ denote the sets of hallucinated and non-hallucinated tokens.  
For a given attention head $h$, we first compute its sensitivity difference:
\begin{equation*}
\mathcal{S}_{h} 
= \frac{1}{|\mathcal{H}|} \sum_{t \in \mathcal{H}} \Delta \mathbb{P}_h(y_t) 
- \frac{1}{|\mathcal{N}|} \sum_{t \in \mathcal{N}} \Delta \mathbb{P}_h(y_t),
\end{equation*}
where
\begin{equation*}
\Delta \mathbb{P}_h(y_t) 
= \mathbb{P}_M(y_t \mid v, x, y_{<t}) 
- \mathbb{P}_{M \setminus h}(y_t \mid v, x, y_{<t})
\end{equation*}
measures the change in the model’s probability of generating token $y_t$ after removing head $h$.  
To further characterize the strength and consistency of this effect, we define a normalized effect size:
\begin{equation*}
\mathcal{E}_{h} = \frac{\mathcal{S}_{h}}
{\sqrt{\sigma_{\mathcal{H}}^2 + \sigma_{\mathcal{N}}^2}},
\end{equation*}
where $\sigma_{\mathcal{H}}^2$ and $\sigma_{\mathcal{N}}^2$ are the variances of $\Delta \mathbb{P}_h(y_t)$ across hallucinated and non-hallucinated tokens, respectively.  
A larger $\mathcal{E}_{h}$ indicates that head $h$ exerts a stronger and more consistent influence on hallucination generation. A smaller $\mathcal{E}_{h}$ indicates that the head has only a weak influence on hallucinations, and removing it causes little to no change in the generation probability of hallucinated tokens.

\subsection{Quantifying the Similarity between Attention Patterns} \label{A.2}
To examine whether hallucination-sensitive heads inherit the attention behaviors of their base language models, we perform a systematic similarity analysis between the attention patterns of Large Vision-Language Models (LVLMs) and its underlying Large Language Models (LLMs). Since LLMs cannot directly process image inputs, we introduce a placeholder token \texttt{<image>} while keeping the textual context identical to ensure consistency. We then focus on the attention patterns of the output tokens.

\paragraph{Extracting Layer--Head Attention Maps.}
For each image--caption pair, we provide both models with the same textual prompt: the LVLM processes the prompt together with the corresponding image, whereas the LLM receives only the textual input through its embedding layer. With \texttt{output\_attentions=True}, both models return full attention matrices
\[
A^{(\ell,h)} \in \mathbb{R}^{T\times T},
\]
where $T$ denotes the number of output tokens.  
Since our comparison concerns output's attention, we extract only the output-related submatrix:
\[
\tilde{A}^{(\ell,h)}
   = A^{(\ell,h)}[-T\!:\,,\,-T\!:].
\]

\paragraph{Similarity Measurement.}
For each layer--head pair, we flatten both attention maps into vectors and compute their cosine similarity:
\[
\mathrm{Sim}^{(\ell,h)}
  = \mathrm{cosine}\!\left(
      \operatorname{vec}\!\left(
        \tilde{A}^{(\ell,h)}_{\mathrm{LVLM}}
      \right),
      \operatorname{vec}\!\left(
        \tilde{A}^{(\ell,h)}_{\mathrm{LLM}}
      \right)
    \right).
\]
This metric directly quantifies how closely the LVLM head reproduces the token-level attention distribution of the LLM.

\paragraph{Hallucination-Sensitive vs. Hallucination-Insensitive Heads.}
We evaluate two type of heads:  
(1) \emph{hallucination-sensitive heads}, which correlate strongly with hallucinated tokens, and  
(2) \emph{hallucination-insensitive heads}, which rarely contribute to hallucinations.  
For each sample, we compute similarity scores for all selected heads and average them across both heads and samples:
\[
\bar{S}_{\mathrm{hal}}
  = \mathbb{E}_{\text{samples}}
      \mathbb{E}_{(\ell,h)\in\mathcal{H}_{\mathrm{hal}}}
        \mathrm{Sim}^{(\ell,h)},\qquad
\bar{S}_{\mathrm{non}}
  = \mathbb{E}_{\text{samples}}
      \mathbb{E}_{(\ell,h)\in\mathcal{H}_{\mathrm{non}}}
        \mathrm{Sim}^{(\ell,h)}.
\]

\section{Theoretical Proofs}
In this section, we formally establish the mathematical formulation linking the query-key parameter matrix \( W_{\mathrm{QK}} \), the attention matrix \( A \), and the attention distribution.

\subsection{Query--Key Parameter Matrix \texorpdfstring{$W_{QK}$}{Wqk} and Attention Distribution}

\subsubsection{Token Propagation Probability}
We adopt the \textit{token propagation probability}~\cite{bao2024self, tang2025intervening} as a quantitative measure of attention localization and uniformity.
To connect it with the attention computation, we begin by approximating the softmax function using a piecewise linear formulation.
For a $T$-dimensional input $\omega \in \mathbb{R}^{T}$, the softmax is defined as
\begin{equation*}
S(\omega)_i = \frac{\exp(\omega_i)}{\sum_{j \in [T]} \exp(\omega_j)}, \quad i \in [T].
\end{equation*}
By performing a first-order Taylor expansion around the origin, we obtain
\begin{equation*}
\gamma^{i} := \nabla_i S(\mathbf{0}) = \frac{1}{T} \mathbf{e}^{i} - \frac{1}{T^2} \mathbf{1}, \quad 
\gamma_0^{i} := S(\mathbf{0})_i = \frac{1}{T}.
\end{equation*}
Then, the piecewise linear approximation of $S(\omega)$ can be expressed as
\begin{equation*}
S(\omega)_i \approx \tilde{S}(\omega)_i = \max \{ 0, \min \{ 1, \langle \gamma^{i}, \omega \rangle + \gamma_0^{i} \} \}.
\end{equation*}
For notational simplicity, we write the vector form as
$\tilde{S}(\omega) = \Gamma^{\top} \omega + \tilde{\gamma}_0$,
where $\Gamma := [\tilde{\gamma}^{1}, \tilde{\gamma}^{2}, \ldots, \tilde{\gamma}^{T}]$ and 
$\tilde{\gamma}_0 := [\tilde{\gamma}_0^{1}, \tilde{\gamma}_0^{2}, \ldots, \tilde{\gamma}_0^{T}]^{\top}$.
Formally, for the $i$-th token, the \textit{signal propagation probability} is defined as
\begin{equation}
\rho_i = \mathbb{P} \left\{ \langle \gamma^{i}, \omega \rangle + \gamma_0^{i} \in [0, 1] \right\},
\end{equation}
where $\omega = \mathbf{X}^\top W_{\mathrm{QK}} \mathbf{X}_T / \sqrt{d}$, and the randomness originates solely from the input tokens $\mathbf{X}$.
When only a few $\rho_i$ take significantly large values, the attention distribution is considered \textit{localized softmax}; 
in contrast, when $\rho_i$ values are similar across tokens, it corresponds to a \textit{uniform softmax}.

\subsubsection{Localized vs. Uniform Attention Distribution}
To analytically characterize $\rho_i$, we follow a synthetic random walk model of token generation:

\noindent\textbf{Assumption 1 (Gaussian Random Walk).}  
The tokens $\mathbf{x}_{t({t \ge 1})}$ are assumed to follow a Gaussian random walk, analogous to the random walk described by Pearson \cite{rawlings2003gaussianwalk}, where the first token $\mathbf{x}_1$ is drawn from a Gaussian distribution with mean 0 and covariance matrix $\Sigma$:
\[
\mathbf{x}_1 \sim \mathcal{N}(0, \Sigma),
\]
and for each subsequent token, the value of $\mathbf{x}_{t+1}$ is drawn from a Gaussian distribution centered around the previous token $\mathbf{x}_t$, with the same covariance structure $\Sigma$:
\[
\mathbf{x}_{t+1} \sim \mathcal{N}(\mathbf{x}_t, \Sigma).
\]
This process models a sequence of tokens that evolve through random steps, each governed by a Gaussian distribution with mean equal to the previous token and fixed covariance $\Sigma$.

Under this assumption, we approximate the expression $\langle \gamma^i, \omega \rangle + \gamma_0^i$ as a Gaussian random variable with mean $\mu^i$ and variance $v^i$:
\[
\langle \gamma^i, \omega \rangle + \gamma_0^i \sim \mathcal{N}(\mu^i, v^i).
\]
This approximation simplifies the distribution of the term to a Gaussian, enabling easier analysis.

To estimate the probability $\rho_i$ associated with this Gaussian random variable, we utilize the cumulative distribution function (CDF) of the normal distribution. Specifically, $\rho_i$ can be approximated as:
\begin{align}
\rho_i \approx \frac{1}{2} \left\{ \mathrm{erf} \!\left( \frac{1 - \mu^i}{\sqrt{2v^i}} \right) + \mathrm{erf} \!\left( \frac{\mu^i}{\sqrt{2v^i}} \right) \right\}.
\end{align}
Here, the error function $\mathrm{erf}(z)$ is related to the CDF of the standard normal distribution, and the quantities $\frac{1 - \mu^i}{\sqrt{2v^i}}$ and $\frac{\mu^i}{\sqrt{2v^i}}$ standardize the values of $\mu^i$ and $v^i$ to yield probabilities within the range [0,1]. This approximation thus leverages the properties of the Gaussian distribution to compute $\rho_i$ efficiently.

\noindent \textbf{Lemma 1 (Moment Formulas for Gaussian Quadratic Forms).}
Let $\mathbf{W} \in \mathbb{R}^{d \times d}$ be a symmetric matrix, and let $\mathbf{a}, \boldsymbol{\mu} \in \mathbb{R}^d$ and $\boldsymbol{\Sigma} \in \mathbb{R}^{d \times d}$ be a covariance matrix. For $\mathbf{x} \sim \mathcal{N}(\boldsymbol{\mu}, \boldsymbol{\Sigma})$, the following moment formulae hold:

\begin{align*}
\mathbb{E}[\mathbf{x}^\top \mathbf{W}\mathbf{x}] &= \mathrm{tr}(\mathbf{W}\boldsymbol{\Sigma}) + \boldsymbol{\mu}^\top \mathbf{W}\boldsymbol{\mu}, \\
\mathbb{E}[\mathbf{x}\mathbf{x}^\top] &= \boldsymbol{\Sigma} + \boldsymbol{\mu}\boldsymbol{\mu}^\top, \\[3pt]
\mathbb{E}[\mathbf{a}^\top \mathbf{W}\mathbf{x}\mathbf{x}^\top \mathbf{W}\mathbf{x}]
&= 2\,\mathbf{a}^\top \mathbf{W}\boldsymbol{\Sigma}\mathbf{W}\boldsymbol{\mu}
   + \mathbf{a}^\top \mathbf{W}\boldsymbol{\mu}\bigl\{\mathrm{tr}(\mathbf{W}\boldsymbol{\Sigma})
   + \boldsymbol{\mu}^\top \mathbf{W}\boldsymbol{\mu}\bigr\}, \\[3pt]
\mathbb{E}[\mathbf{x}^\top \mathbf{W}\mathbf{x}\mathbf{x}^\top \mathbf{W}\mathbf{x}]
&= 2\,\mathrm{tr}(\mathbf{W}\boldsymbol{\Sigma}\mathbf{W}\boldsymbol{\Sigma})
   + \{\mathrm{tr}(\mathbf{W}\boldsymbol{\Sigma})\}^2
   + 4\,\boldsymbol{\mu}^\top \mathbf{W}\boldsymbol{\Sigma}\mathbf{W}\boldsymbol{\mu} \notag\\
&\quad + 2\,\mathrm{tr}(\mathbf{W}\boldsymbol{\Sigma})\,\boldsymbol{\mu}^\top \mathbf{W}\boldsymbol{\mu}
   + \boldsymbol{\mu}^\top \mathbf{W}\boldsymbol{\mu}\,\boldsymbol{\mu}^\top \mathbf{W}\boldsymbol{\mu}. 
\end{align*}

For $i \le j$, suppose that $\mathbf{x}_i, \mathbf{x}_j$ follow Assumption~1. Then, the following formulae hold:

\begin{align*}
\mathbb{E}[\mathbf{x}_i^\top \mathbf{W}\mathbf{x}_i]
&= (i - 1)\,\mathrm{tr}(\mathbf{W}\boldsymbol{\Sigma}), \\
\mathbb{E}[\mathbf{x}_i^\top \mathbf{W}\mathbf{x}_i \mathbf{x}_i^\top \mathbf{W}\mathbf{x}_i]
&= (i^2 - 2i + 2)\bigl\{2\,\mathrm{tr}(\mathbf{W}\boldsymbol{\Sigma}\mathbf{W}\boldsymbol{\Sigma})
   + \{\mathrm{tr}(\mathbf{W}\boldsymbol{\Sigma})\}^2\bigr\}, \\
\mathbb{E}[\mathbf{x}_i^\top \mathbf{W}\mathbf{x}_i \mathbf{x}_j^\top \mathbf{W}\mathbf{x}_j]
&= (i^2 + ij - 3i - j + 4)\,\mathrm{tr}(\mathbf{W}\boldsymbol{\Sigma}\mathbf{W}\boldsymbol{\Sigma})
   + (i^2 - 2i + 2)\{\mathrm{tr}(\mathbf{W}\boldsymbol{\Sigma})\}^2, \\
\mathbb{E}[\mathbf{x}_i^\top \mathbf{W}\mathbf{x}_j \mathbf{x}_j^\top \mathbf{W}\mathbf{x}_j]
&= (ij - i - j + 2)\bigl\{2\,\mathrm{tr}(\mathbf{W}\boldsymbol{\Sigma}\mathbf{W}\boldsymbol{\Sigma})
   + \{\mathrm{tr}(\mathbf{W}\boldsymbol{\Sigma})\}^2\bigr\}. 
\end{align*}

The formulas in Lemma~1 are standard and can also be found in prior studies \cite{brookes1998matrix, bao2024self}.

\noindent \textbf{Lemma 2.}
Suppose that $\mathbf{W}_{QK}$ is symmetric (for the asymmetric case, the probability is calculated using the symmetrized matrix \( \frac{1}{2}(W_{\mathrm{QK}} + W_{\mathrm{QK}}^{\top}) \)) and independent from $\mathbf{X}$, and let $\mathbf{W} := \mathbf{W}_{QK}\Sigma$. 
Under Assumption~1, for $i \in [T]$, the mean $\mu^i$ and variance $v^i$ of 
$\langle \gamma^i, \omega \rangle + \gamma_0^i$ with the input $\omega := \mathbf{X}^\top \mathbf{W}_{QK}\mathbf{x}_T / \sqrt{d}$
are given as follows:
\begin{align}
\mu^i &= \left( \frac{i}{T} - \frac{1}{2} \right) \frac{\mathrm{tr}(\mathbf{W})}{\sqrt{d}} + o(1), \\
v^i &= \left( \frac{2i^2}{T^2} + \frac{7}{12} \right) \frac{\mathrm{tr}(\mathbf{W}^2)}{d} + o(1).
\end{align}

\noindent\textit{Proofs.} We use Lemma 1 to derive the $\mu^i$:
\begin{align*}
\mu^i 
&= \frac{1}{\sqrt{d} T}\mathbb{E}[\mathbf{x}_i^\top \mathbf{W}_{QK}\mathbf{x}_T]
    - \frac{1}{\sqrt{d} T^2}\sum_{j\in[T]}\mathbb{E}[\mathbf{x}_j^\top \mathbf{W}_{QK}\mathbf{x}_T] + o(1) \\
&= \frac{i-1}{\sqrt{d} T}\mathrm{tr}(\mathbf{W}) 
    - \frac{\sum_{j\in[T]}(j-1)}{\sqrt{d} T^2}\mathrm{tr}(\mathbf{W}) + o(1) \\
&= \left( \frac{2i-T+1}{2\sqrt{d}T} \right) {\mathrm{tr}(\mathbf{W})} + o(1).
\end{align*}

\noindent
To derive the variance, we first evaluate the expectation
$\mathbb{E}[\mathbf{x}_i^\top \mathbf{W}_{QK}\mathbf{x}_T \mathbf{x}_j^\top \mathbf{W}_{QK}\mathbf{x}_T]$ 
(for $i \le j \le T$):
\small\begin{align*}
\mathbb{E}[\mathbf{x}_i^\top \mathbf{W}_{QK}\mathbf{x}_T \mathbf{x}_j^\top \mathbf{W}_{QK}\mathbf{x}_T] 
&= \mathbb{E}[\mathbf{x}_i^\top \mathbf{W}_{QK} (\mathbf{x}_T \mathbf{x}_T^\top)\mathbf{W}_{QK}\mathbf{x}_j] \\
&= \mathbb{E}[\mathbf{x}_i^\top \mathbf{W}_{QK}((T-j)\Sigma + \mathbf{x}_j \mathbf{x}_j^\top)\mathbf{W}_{QK}\mathbf{x}_j] \\
&= (T-j)\mathbb{E}[\mathbf{x}_i^\top \mathbf{W}_{QK}\Sigma \mathbf{W}_{QK}\mathbf{x}_j]
   + \mathbb{E}[\mathbf{x}_i^\top \mathbf{W}_{QK}\mathbf{x}_j \mathbf{x}_j^\top \mathbf{W}_{QK}\mathbf{x}_j] \\
&= (T-j)(i-1)\mathrm{tr}(\mathbf{W}^2)
   + (ij - i - j + 2)\{2\mathrm{tr}(\mathbf{W}^2)+\mathrm{tr}(\mathbf{W})^2\} \\
&= [(i - 1)(T + j - 2) + 2\bigr]\,\mathrm{tr}(\mathbf{W}^2)
+ [(i - 1)(j - 1) + 1]\,\mathrm{tr}(\mathbf{W})^2.
\end{align*}

Then, the expectation of the squared term is expanded:
\begin{align*}
\mathbb{E}\!\left[\langle \gamma^i, \mathbf{X}^\top \mathbf{W}_{QK}\mathbf{x}_T \rangle^2\right]
&= \mathbb{E}\!\left[\left(\frac{1}{T}\mathbf{x}_i^\top \mathbf{W}_{QK}\mathbf{x}_T
   - \frac{1}{T^2}\sum_{j\in[T]} \mathbf{x}_j^\top \mathbf{W}_{QK}\mathbf{x}_T\right)^2\right] \\
&= \frac{1}{T^2}\mathbb{E}[\mathbf{x}_i^\top \mathbf{W}_{QK}\mathbf{x}_T \mathbf{x}_i^\top \mathbf{W}_{QK}\mathbf{x}_T]
 - \frac{2}{T^3}\sum_{j\in[T]}\mathbb{E}[\mathbf{x}_i^\top \mathbf{W}_{QK}\mathbf{x}_T \mathbf{x}_j^\top \mathbf{W}_{QK}\mathbf{x}_T] \nonumber\\
&\quad + \frac{1}{T^4}\sum_{j,j'\in[T]}\mathbb{E}[\mathbf{x}_j^\top \mathbf{W}_{QK}\mathbf{x}_T \mathbf{x}_{j'}^\top \mathbf{W}_{QK}\mathbf{x}_T].
\end{align*}

By simplifying the equations, we obtain:
\begin{align*}
\mathbb{E}\!\left[\langle \gamma^i, \mathbf{X}^\top \mathbf{W}_{QK}\mathbf{x}_T \rangle^2\right]
= \left(\frac{7}{12} + \frac{2i^2}{T^2}\right)\mathrm{tr}(\mathbf{W}^2)
  + \left(\frac{1}{4} - \frac{i}{T} + \frac{i^2}{T^2}\right)\mathrm{tr}(\mathbf{W})^2 + o(1).
\end{align*}

We derived the $v^i$:
\begin{align*}
v^i 
&= \mathbb{V}[\langle \gamma^i, \omega \rangle] \\
&= \frac{1}{d}\left[\mathbb{E}\!\left[\langle \gamma^i, \mathbf{X}^\top \mathbf{W}_{QK}\mathbf{x}_T\rangle^2\right]
   - (\mu^i)^2\right] \\
&= \frac{1}{d}\left(\frac{7}{12} + \frac{2i^2}{T^2}\right)\mathrm{tr}(\mathbf{W}^2) + o(1).
\end{align*}



By extending $\mu^i$ and $v^i$ smoothly over $\theta = i/T \in [0, 1]$, 
we define the \textit{token propagation probability} $\rho(\theta)$ as

\small
\begin{equation}
\rho(\theta)
= \Phi\!\left( \left(\theta - \tfrac{1}{2}\right)\xi; \theta \right)
- \Phi\!\left( \left(\theta - \tfrac{1}{2}\right)\xi - \tfrac{1}{\eta}, \theta \right),
\end{equation}
\normalsize
where
\begin{align*}
\xi &= \frac{\mathrm{tr}(\mathbf{W})}{\sqrt{\mathrm{tr}(\mathbf{W}^2)}}, 
\quad
\eta = \frac{\sqrt{\mathrm{tr}(\mathbf{W}^2)}}{\sqrt{d}}, \\
\Phi(z; \theta) &= 
\frac{1}{2}\!\left[   
\mathrm{erf}\!\!\left(
\frac{z}{\sqrt{(4\theta^2 + \tfrac{7}{6})}}
\right)\right].
\end{align*}

Substituting the expression for $\Phi(z;\theta)$ into the definition of $\rho(\theta)$ gives
\begin{equation}
\rho(\theta)
= \frac{1}{2}\!\left[
\mathrm{erf}\!\left(
\frac{(\theta-\tfrac{1}{2})\xi}{\sqrt{4\theta^2+\tfrac{7}{6}}}
\right)
-
\mathrm{erf}\!\left(
\frac{(\theta-\tfrac{1}{2})\xi-\tfrac{1}{\eta}}{\sqrt{4\theta^2+\tfrac{7}{6}}}
\right)
\right].
\end{equation}
Next, by substituting $\xi$ and $\eta$, we obtain the compact closed-form expression:
\begin{equation}
\rho(\theta)
= \tfrac{1}{2}\!\left[
\mathrm{erf}\!\left(
\frac{(\theta-\tfrac{1}{2})\,\mathrm{tr}(\mathbf{W})}
{\sqrt{\mathrm{tr}(\mathbf{W}^2)}\,\sqrt{4\theta^2+\tfrac{7}{6}}}
\right)
-
\mathrm{erf}\!\left(
\frac{(\theta-\tfrac{1}{2})\,\mathrm{tr}(\mathbf{W})-\sqrt{d}}
{\sqrt{\mathrm{tr}(\mathbf{W}^2)}\,\sqrt{4\theta^2+\tfrac{7}{6}}}
\right)
\right].
\end{equation}

\noindent \textbf{Results.}
From the above derivation, the behavior of $\rho(\theta)$ is governed by the trace terms 
$\mathrm{tr}(\mathbf{W})$ and $\mathrm{tr}(\mathbf{W}^2)$, 
which correspond to the first- and second-order moments of the eigenvalue spectrum of $W_{\mathrm{QK}}$.
Specifically:
\begin{itemize}
    \item \textbf{Localized regime.} 
    $\rho(\theta)$ exhibits a peaked (localized) pattern when 
    $\lvert \mathrm{tr}(\mathbf{W}) \rvert \gtrsim \sqrt{d}$ 
    so that the peak position 
    $\theta^\star = \tfrac{1}{2} + \tfrac{\sqrt{d}}{2\,\mathrm{tr}(\mathbf{W})}$ 
    lies within $(0,1)$, 
    and when $\mathrm{tr}(\mathbf{W}^2)$ is close to zero. 
    In this case, the attention distribution shows clear concentration.

    \item \textbf{Uniform regime.} 
    When $\mathrm{tr}(\mathbf{W})$ is close to zero 
    and $\mathrm{tr}(\mathbf{W}^2)$ is finite, 
    the function $\rho(\theta)$ becomes smooth without a directional bias,
    indicating an approximately uniform attention distribution.

\end{itemize}

\subsection{Attention Matrix \texorpdfstring{$A$}{A} and Attention Distribution}

Given the query–key interaction $QK^{\top}$, the attention matrix is defined as
\begin{equation*}
A = \mathrm{softmax}\!\left(\frac{QK^{\top}}{\sqrt{d}}\right),
\end{equation*}
where each row $A_i = (A_{i1}, A_{i2}, \dots, A_{iT})$ represents the attention distribution of token $i$ over all contextual tokens. 
The statistical property of $A_i$ reflects how the model allocates its focus across the input sequence.

\paragraph{Variance as a measure of concentration.}
To quantify the concentration of attention, we compute the row-wise variance of $A$:
\begin{equation*}
\sigma_A^2 = \frac{1}{T} \sum_{j=1}^{T} (A_{ij} - \tfrac{1}{T})^2.
\end{equation*}
A small $\sigma_A^2$ indicates a nearly uniform attention distribution, where each token receives comparable importance.
Conversely, a large $\sigma_A^2$ implies that the attention weights are sharply peaked, signaling localized or selective attention.
This variance measure serves as a simple yet effective descriptor of the attention distribution’s shape.

\paragraph{Connection to entropy.}
The variance of $A_i$ is inversely related to its entropy $H(A_i) = -\sum_j A_{ij}\log A_{ij}$:
\begin{equation*}
H(A_i) \uparrow \;\Leftrightarrow\; \sigma_A^2 \downarrow,
\qquad
H(A_i) \downarrow \;\Leftrightarrow\; \sigma_A^2 \uparrow.
\end{equation*}
This negative correlation reflects a fundamental principle from information theory: 
for a discrete distribution over a fixed support, the entropy reaches its maximum when the distribution is uniform and decreases as the distribution becomes more concentrated 
\citep{cover2006elements, jaynes1957information}. 
Accordingly, the variance serves as a complementary descriptor of attention entropy, providing an analytic proxy for quantifying how ``localized'' or ``uniform'' the attention distribution is.



\section{Experimental Details}

\subsection{Detailed Descriptions and Experimental Setup of Baselines}
We provide detailed descriptions and experimental settings of the baseline methods used for comparison in our experiments.

\paragraph{FarSight \cite{tang2025seeing}} 
FarSight is a decoding-based hallucination mitigation method designed to enhance information propagation during generation. It constructs causal masks to reinforce effective interactions among multimodal tokens and to suppress attention drift toward outlier tokens. By dynamically adjusting the causal mask through an attention-register structure and a positional-aware masking mechanism, FarSight effectively reduces both initial hallucinations and snowball hallucinations across image and video tasks.

\paragraph{VCD \cite{10657718}}
Visual Contrastive Decoding (VCD) mitigates object hallucination by introducing contrastive visual signals during decoding. 
Specifically, VCD constructs a positive image (the original input) and a negative image (a visually perturbed version), and contrasts the logits produced under these two conditions. 
Tokens whose logits do not sufficiently depend on the true visual input are down-weighted, reducing visually ungrounded generations. 
VCD is lightweight and model-agnostic, requiring no training or model modification.

\paragraph{DoLA \cite{chuang2024dola}}
Decoding by Contrasting Layers (DoLa) improves factuality in LLMs by contrasting the logits of deep layers (which are more factual) with shallow layers (which may contain stylistic or biased information). 
The final logits are obtained by subtracting the shallow-layer distribution from the deep-layer distribution. 
Although originally proposed for LLMs, DoLa can be applied to LVLMs by performing layer-wise logit contrast during multimodal decoding, thus reducing hallucinated or unsupported content.

\paragraph{HALC \cite{pmlr-v235-chen24bi}}
Hallucination Reduction via Adaptive Focal-Contrast Decoding (HALC) is a decoding method designed specifically for object hallucination in LVLMs. 
It applies a focal-style modulation to down-weight uncertain or weakly grounded visual tokens, and introduces a contrastive correction that penalizes over-confident hallucinations. 
HALC adaptively adjusts the correction strength based on token-level confidence, enabling fine-grained control over hallucination mitigation without harming general performance.

\paragraph{OPERA \cite{DBLP:conf/cvpr/HuangDZ0H0L0Y24}}
Over-trust Penalty and a Retrospection-Allocation strategy (OPERA) reduces hallucination via two key mechanisms: \emph{Over-Trust Penalty} and \emph{Retrospection-Allocation}. 
The first penalizes the model when its textual prediction becomes overly confident in the absence of sufficient visual grounding. 
The second reallocates attention across vision tokens by retrospectively checking whether generated tokens are visually supported. 
Together, these mechanisms enforce alignment between the model’s confidence and its visual evidence.

\paragraph{AD-HH \cite{DBLP:conf/iclr/YangL0X25}}
AD-HH analyzes hallucination sources in LVLMs via causal mediation analysis and identifies specific hallucination heads within Multi-Head Attention as key contributors. It introduces a training-free decoding intervention and a lightweight fine-tuning strategy that reduce the over-reliance of these heads on text tokens, achieving substantial improvements in hallucination mitigation across captioning benchmarks.

\paragraph{Experimental Setup}
For a fair and consistent comparison, we adopt the official configurations and recommended hyperparameters for all baseline methods. For \textbf{FarSight}, we set the decay rate $\sigma$ to $\log_{\alpha}(\mathrm{seq})$, where $\alpha = 1024$ denotes the typical maximum token limit. \textbf{DoLA} is implemented with an adaptive plausibility threshold of $0.1$ and an early-exit schedule applied to layers $[0, 2, 4, \dots, 32]$. \textbf{OPERA} uses a self-attention scaling factor of $50$, an attending-retrospection threshold of $15$, a beam size of $5$, and a penalty weight of $1.0$. For \textbf{VCD}, we follow the standard configuration with an amplification factor of $1$, an adaptive plausibility threshold of $0.1$, and $500$ diffusion noise steps. \textbf{HALC} is implemented with the official DINO-based detector; we set the JSD buffer size to $m = 6$, beam size to $1$, number of sampled fields-of-view to $n = 4$, exponential growth ratio of contextual fields to $0.6$, bounding-box threshold to $0.4$, and adaptive plausibility threshold to $0.1$. For \textbf{AD-HH}, we reweight the top $20$ attention heads and use $0.5$ as the reweighting threshold.

\subsection{Hyperparameter Sensitivity Analysis}

\begin{figure}[t]
	  \centering
	  \includegraphics[width=1\textwidth]{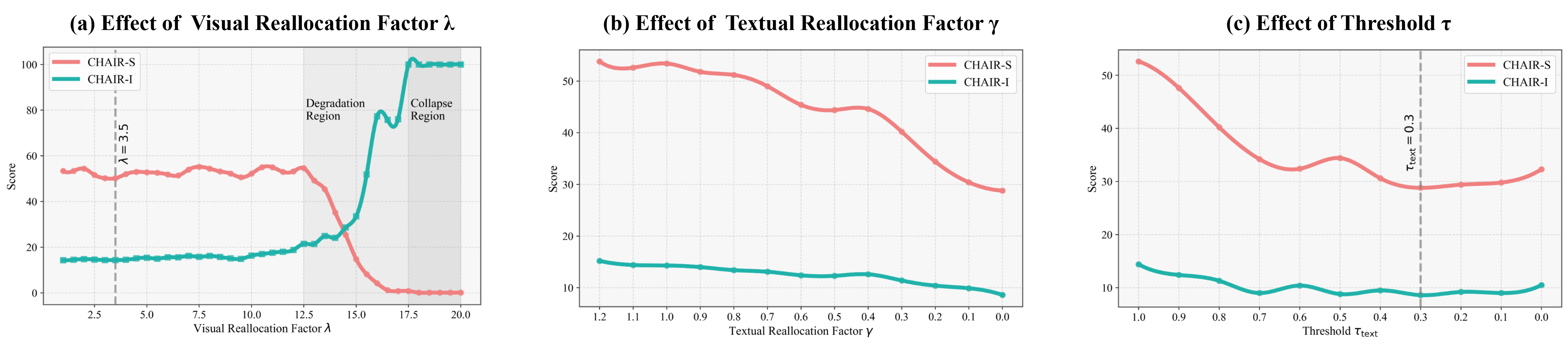}
    \caption{Sensitivity of the attention reallocation mechanism to hyperparameters. (a) shows the impact of the visual reallocation factor $\lambda$. (b) illustrates the effect of the textual reallocation factor $\gamma$. (c) demonstrates the influence of the attention reallocation threshold $\tau_{\mathrm{text}}$.}
	  \label{fig5}
\end{figure}

\noindent\textbf{Effect of the visual reallocation factors $\lambda$.}
Figure \ref{fig5} (a) illustrates the effects of the visual reallocation factor $\lambda$ on hallucination mitigation.
For the factor $\lambda$, a moderate increase enhances the model’s attention to visual information during decoding, thereby effectively reducing hallucinations. The model achieves its best performance at $\lambda = 3.5$. However, once $\lambda$ exceeds the threshold of 12.5, the performance begins to deteriorate noticeably, and when $\lambda > 17.5$, the model collapses entirely, producing repetitive outputs. 

\noindent\textbf{Effect of the textual reallocation factors $\gamma$.}
Figure \ref{fig5} (b) illustrates the effects of the textual reallocation factor $\gamma$ on hallucination mitigation.
As $\gamma$ decreases, we observe a gradual reduction in object hallucinations. Since reallocation is applied only to hallucination-sensitive attention heads rather than all attention heads, setting $\gamma$ to a lower value does not lead to abnormal model behavior. On the contrary, by suppressing excessive attention to the text, the modality-wise attention imbalance (MAI) across the average attention heads is reduced, achieving a better balance.

\noindent\textbf{Effect of the attention reallocation threshold $\tau_{\mathrm{text}}$.}
Figure \ref{fig5} (c) shows how the attention reallocation threshold $\tau_{\mathrm{text}}$ influences the model’s behavior. As the threshold decreases from 1.0, both \(C_S\) and \(C_I\) scores show an overall downward trend, indicating that a lower threshold helps trigger attention reallocation earlier and suppress hallucinations caused by excessive focus on textual modalities. The model achieves optimal performance when $\tau_{\mathrm{text}}$ achieves 0.3. However, further decreasing the threshold leads to overly frequent reallocation, introducing noise perturbations and causing slight performance fluctuations.

\noindent\textbf{Effect of the regularization coefficient $\beta$.}
Table \ref{tab:air_two_module_ablation1} reports the effect of different regularization coefficients $\beta$. As $\beta$ increases from 0 to 0.3, all metrics improve notably, indicating that moderate mean-shift regularization helps stabilize the attention distribution and suppress hallucinations. When $\beta$ is raised to 0.6, the model shows a slight performance drop but remains close to the optimal setting. In contrast, further increasing $\beta$ to 0.9 leads to excessive homogenization, weakening the expressive capacity of attention and resulting in clear degradation on MM-Vet. Overall, a moderate level of regularization (e.g., $\beta = 0.3\text{--}0.6$) proves most effective.

\begin{table}[ht]
	\centering
	\caption{Effect of the regularization coefficient $\beta$.}
	\label{tab:air_two_module_ablation1}
	\setlength{\tabcolsep}{8pt}
	\footnotesize
	\begin{tabular}{lcccc}
		\toprule
		&  \(C_S\!\downarrow\) &  \(C_I\!\downarrow\) & \(F_1\!\uparrow\) & \textbf{MM-Vet Overall}\,$\uparrow$\\
		\midrule
		$\beta=0$   &32.1 &9.9  &0.85  &30.5   \\
		$\beta=0.3$     &\textbf{28.8}  &\textbf{8.6}  &\textbf{0.86}  &\textbf{32.0} (\textbf{+4.9\%})   \\
		$\beta=0.6$     &30.4  &9.0&0.86&31.9 (+4.6\%)\\
		$\beta=0.9$      & 32.0  &9.2&0.85&27.4 (-10.2\%)\\
		\bottomrule
	\end{tabular}
\end{table}

\subsection{Additional Experimental Results}

\begin{figure}[ht]
	  \centering
	  \includegraphics[width=0.6\textwidth]{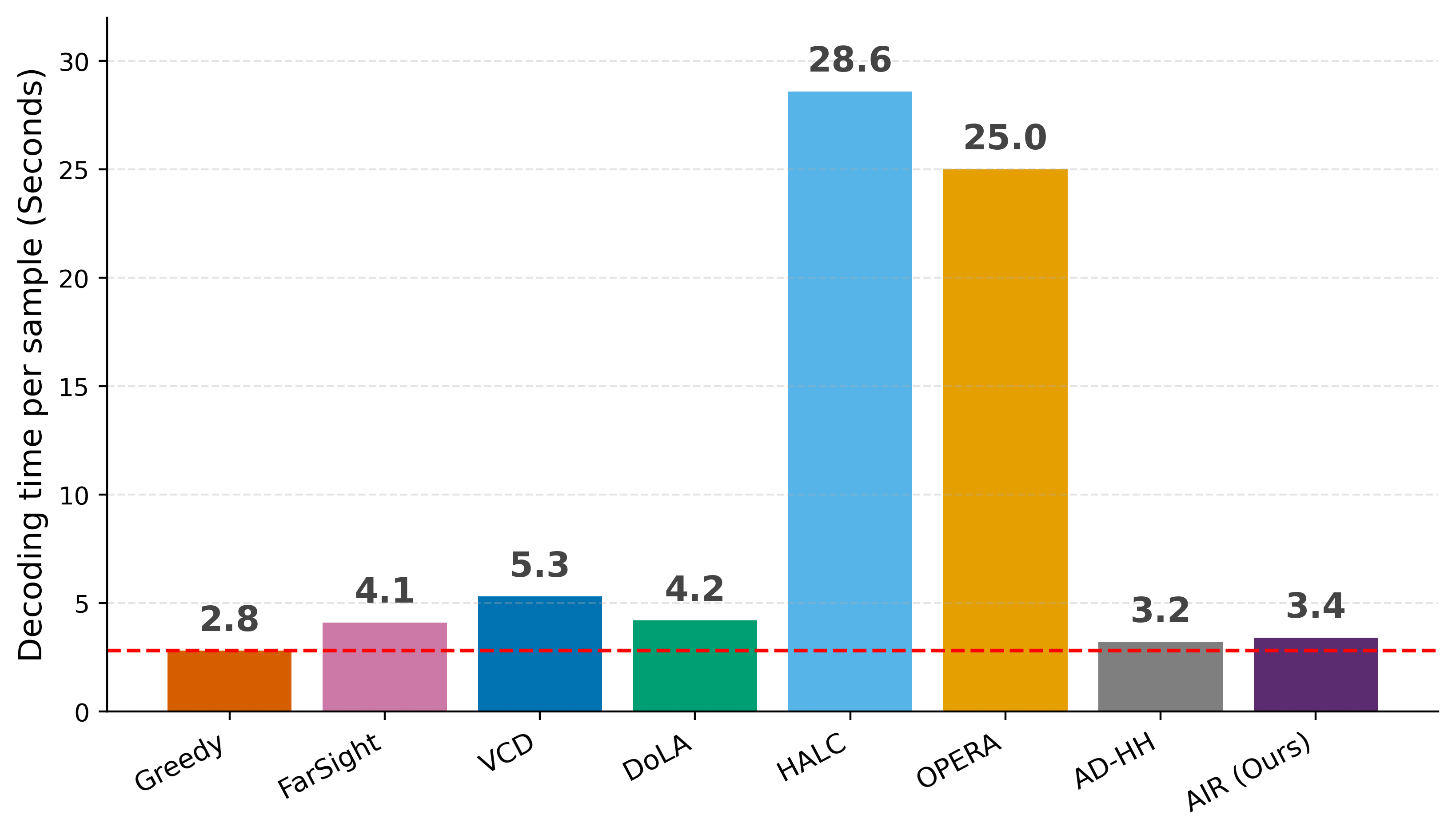}
    \caption{Comparison of inference time across baselines. Except for HALC and OPERA, most methods maintain an average decoding time of under 6 seconds per sample.}
	  \label{time}
\end{figure}

\noindent\textbf{Comparison of generation time.}
Figure~\ref{time} presents the average decoding time per sample for all compared methods. Greedy decoding provides the fastest speed at 2.8 seconds. AIR achieves a similarly efficient latency of \textbf{3.4} seconds, which is close to AD-HH at 3.2 seconds. Other lightweight baselines, including FarSight (4.1 seconds), DoLA (4.2 seconds), and VCD (5.3 seconds), also remain below the 6-second range. In contrast, HALC and OPERA introduce substantial computational cost, requiring 28.6 seconds and 25.0 seconds per sample respectively. These results indicate that AIR offers competitive efficiency while maintaining strong performance in hallucination mitigation.

\noindent\textbf{Evaluation of additional LVLM models.} 
We further evaluate AIR on four additional LVLMs to assess its generalizability across different architectures. As shown in Table~\ref{tab:chair_results1}, AIR consistently lowers both the object-level ($C_S$) and image-level ($C_I$) hallucination scores compared with greedy decoding. The improvements are substantial across all evaluated models, demonstrating that AIR is not tied to a specific backbone. In particular, the largest relative reductions are observed on DeepSeek-VL2-4.5B, where hallucination rates decrease by around 40\% on both metrics. Similar gains are also achieved on LLaVA-NeXT-7B, Qwen-2.5-VL-7B, and InternVL-3.5-8B, indicating that AIR effectively mitigates hallucinations across models with distinct training paradigms and visual encoders. These results highlight the robustness of AIR for improving the reliability of LVLM-generated responses.


\noindent\textbf{AIR’s performance across different scales of LVLMs.}
Across LLaVA models of varying capacities (7B, 13B, and 34B), AIR consistently reduces hallucination rates under the CHAIR metric. As shown in Table~\ref{scale}, AIR delivers substantial improvements over the greedy decoding baseline, achieving up to \textbf{44.4\%} and \textbf{37.2\%} relative reductions in \(C_S\) and \(C_I\) on the 7B model. The improvements remain robust as model size increases: AIR reduces hallucinations by \textbf{33.0\%} and \textbf{28.0\%} on the 13B model, and continues to provide measurable benefits on the stronger 34B LLaVA-NeXT model. These results indicate that AIR scales effectively with model capacity and remains beneficial even for large LVLMs that already exhibit lower hallucination levels.

\begin{table*}[t]
    \centering
    \caption{\textbf{CHAIR} hallucination evaluation results on additional four LVLMs across baselines. Smaller \(C_S\) and \(C_I\) indicate fewer hallucinations. Results are reported with maximum new tokens set to 256. $\Delta\%$ represents the relative improvement over the baseline. Our method AIR consistently enhances hallucination mitigation across all additional LVLMs.}
    \label{tab:chair_results1}
    \setlength{\tabcolsep}{5pt}
    \footnotesize
    \begin{threeparttable}
        \begin{tabular}{lcccccccc}
            \toprule
            \rowcolor{gray!10}
            & \multicolumn{8}{c}{\textbf{Max New Tokens: 256}} \\
            \cmidrule(lr){1-9}
            \multirow{2}{*}{\textbf{Methods}}
            & \multicolumn{2}{c}{LLaVA-NeXT-7B \cite{li2024llava}} 
            & \multicolumn{2}{c}{Qwen-2.5‑VL-7B \cite{bai2025qwen25vltechnicalreport}}
            & \multicolumn{2}{c}{InternVL-3.5-8B \cite{chen2024internvl}}
            & \multicolumn{2}{c}{DeepSeek-VL2-4.5B \cite{wu2024deepseekvl2mixtureofexpertsvisionlanguagemodels}} \\
            & \(C_S\!\downarrow\) & \(C_I\!\downarrow\)
            & \(C_S\!\downarrow\) & \(C_I\!\downarrow\)
            & \(C_S\!\downarrow\) & \(C_I\!\downarrow\)
            & \(C_S\!\downarrow\) & \(C_I\!\downarrow\) \\
            \midrule

            Greedy   &31.2  &8.6  & 38.0 & 9.4 &32.9 & 9.0 & 42.4 & 11.6 \\

            \rowcolor{gray!10}
            \textbf{AIR (Ours)} 
            & \textbf{22.5} & \textbf{6.1}
            & \textbf{29.3} & \textbf{8.8}
            & \textbf{25.1} & \textbf{8.3}
            & \textbf{25.8} & \textbf{6.9} \\

            $\Delta\%$
            & \textcolor{Blue}{$\downarrow$27.9\%} & \textcolor{Blue}{$\downarrow$29.1\%}
            & \textcolor{Blue}{$\downarrow$22.9\%} & \textcolor{Blue}{$\downarrow$6.4\%}
            & \textcolor{Blue}{$\downarrow$23.7\%} & \textcolor{Blue}{$\downarrow$7.8\%}
            & \textcolor{Blue}{$\downarrow$39.2\%} & \textcolor{Blue}{$\downarrow$40.5\%} \\

            \bottomrule
        \end{tabular}
    \end{threeparttable}
\end{table*}

\begin{table*}[t]
    \centering
    \caption{\textbf{CHAIR} hallucination evaluation results across LVLMs of different scales. Smaller \(C_S\) and \(C_I\) indicate fewer hallucinations. All results are reported with the maximum new token limit set to 256. $\Delta\%$ denotes the relative improvement over the baseline. Our method AIR consistently improves hallucination mitigation across models of varying scales.}
    \label{scale}
    \setlength{\tabcolsep}{6pt}
    \footnotesize
    \begin{threeparttable}
        \begin{tabular}{lcccccc}
            \toprule
            \rowcolor{gray!10}
            & \multicolumn{6}{c}{\textbf{Max New Tokens: 256}} \\
            \cmidrule(lr){1-7}
            \multirow{2}{*}{\textbf{Methods}}
            & \multicolumn{2}{c}{LLaVA-1.5-7B} 
            & \multicolumn{2}{c}{LLaVA-1.5-13B}
            & \multicolumn{2}{c}{LLaVA-NeXT-34B} \\
            & \(C_S\!\downarrow\) & \(C_I\!\downarrow\)
            & \(C_S\!\downarrow\) & \(C_I\!\downarrow\)
            & \(C_S\!\downarrow\) & \(C_I\!\downarrow\) \\
            \midrule
            Greedy   & 51.8 & 13.7 & 46.3 & 12.5 & 25.7 & 6.1 \\
            \rowcolor{gray!10}
            \textbf{AIR (Ours)} 
            & \textbf{28.8} & \textbf{8.6}
            & \textbf{31.0} & \textbf{9.0}
            & \textbf{23.2} & \textbf{6.0} \\
            $\Delta\%$
            & \textcolor{Blue}{$\downarrow$44.4\%} & \textcolor{Blue}{$\downarrow$37.2\%}
            & \textcolor{Blue}{$\downarrow$33.0\%} & \textcolor{Blue}{$\downarrow$28.0\%}
            & \textcolor{Blue}{$\downarrow$9.7\%} & \textcolor{Blue}{$\downarrow$1.6\%} \\
            \bottomrule
        \end{tabular}
    \end{threeparttable}
\end{table*}

\begin{figure}[ht]
	  \centering
	  \includegraphics[width=1\textwidth]{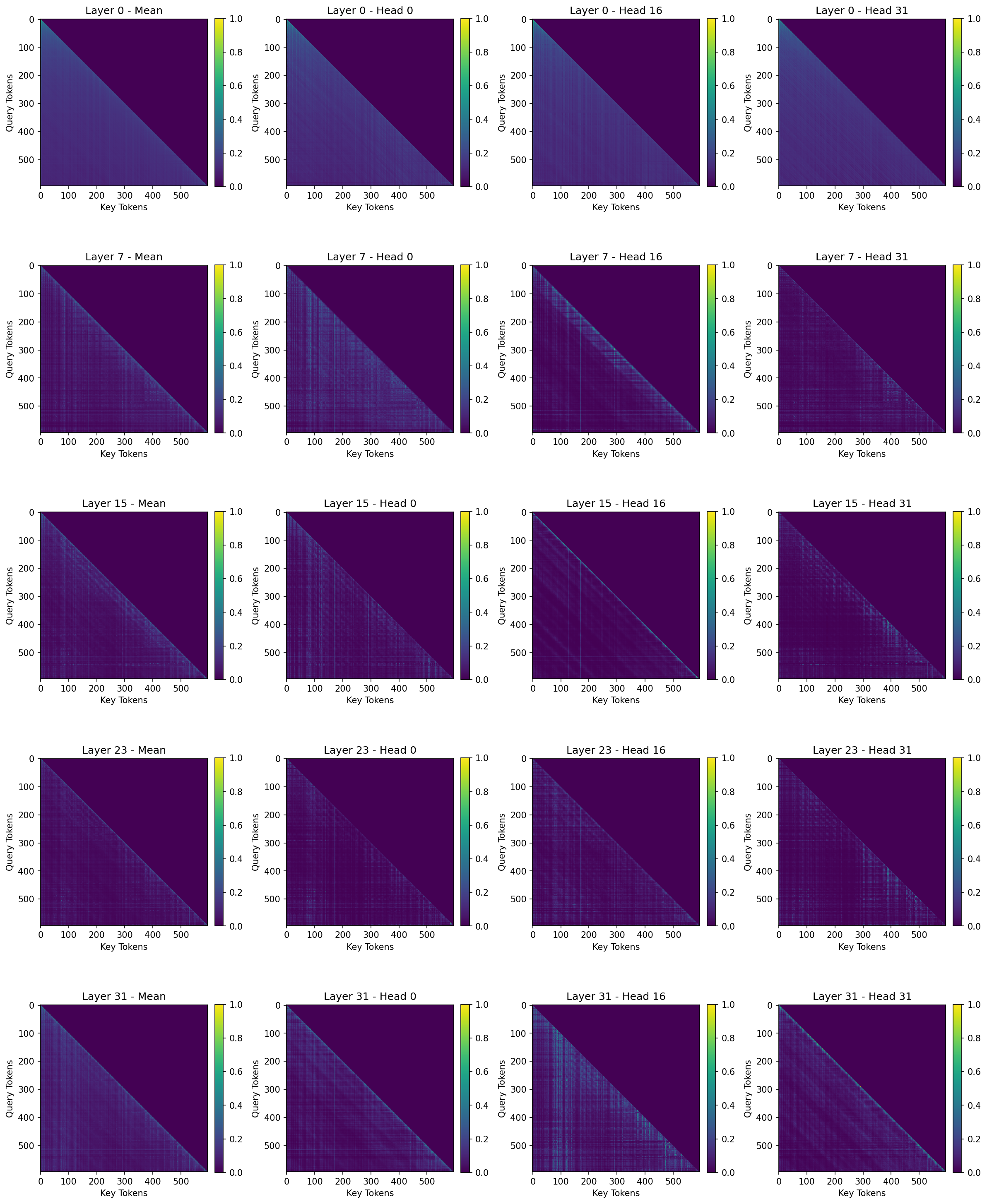}
    \caption{Attention maps across layers and heads for Case One.}
	  \label{case1}
\end{figure}

\begin{figure}[ht]
	  \centering
	  \includegraphics[width=1\textwidth]{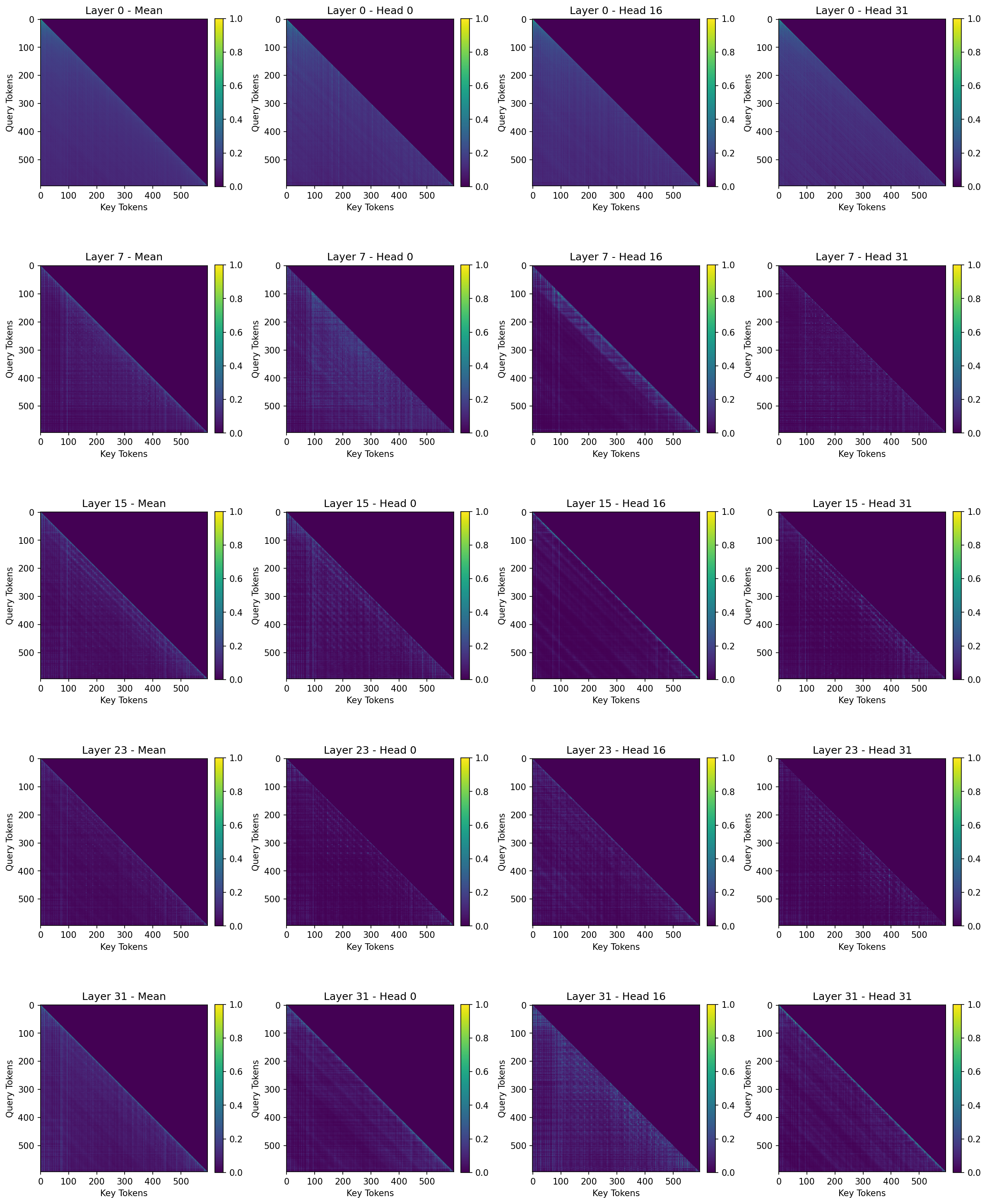}
    \caption{Attention maps across layers and heads for Case Two.}
	  \label{case2}
\end{figure}

\begin{figure}[ht]
	  \centering
	  \includegraphics[width=1\textwidth]{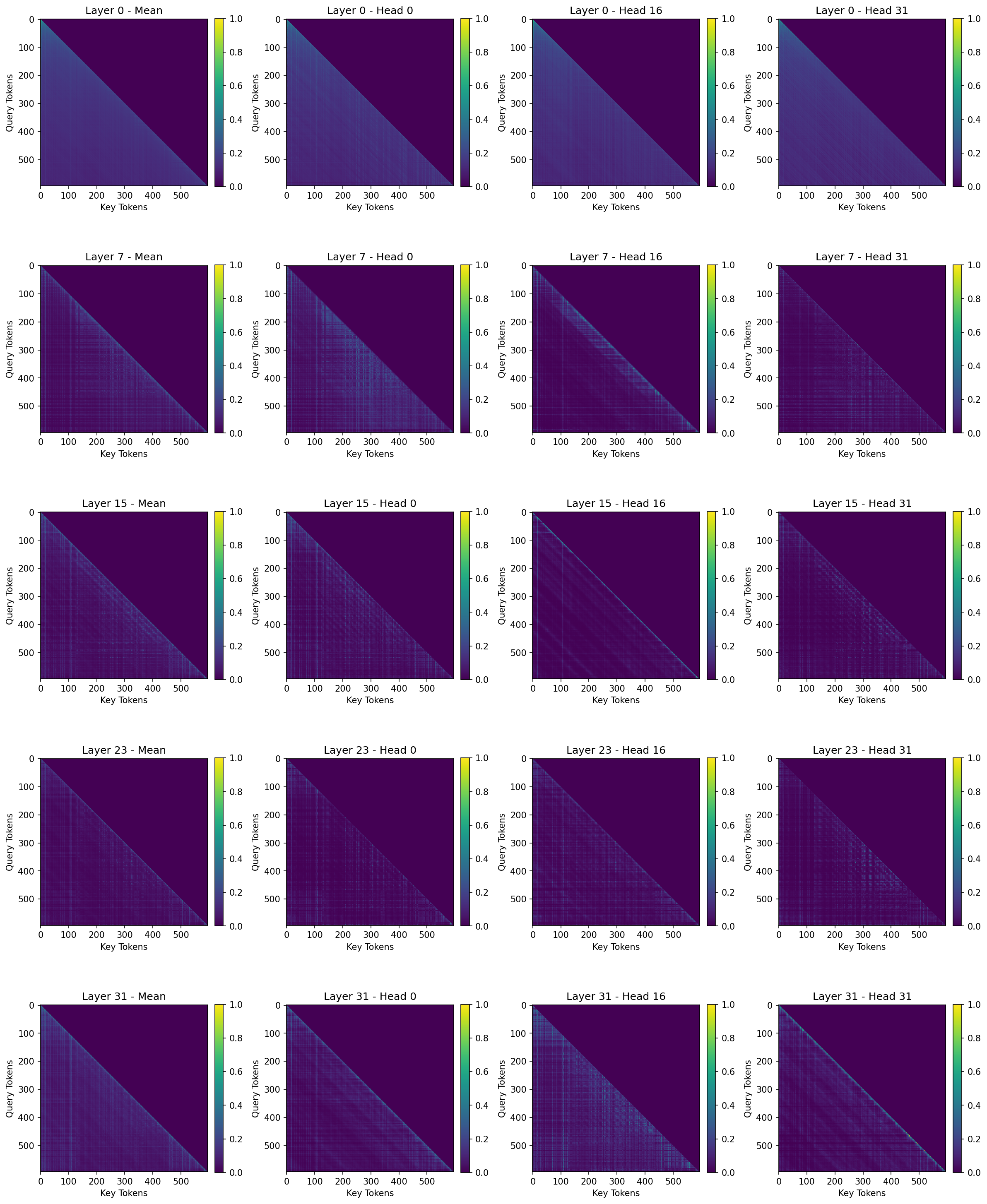}
    \caption{Attention maps across layers and heads for Case Three.}
	  \label{case3}
\end{figure}

\noindent\textbf{Visual analysis of TAI across layers and attention heads.}
We visualize attention maps under hallucination-generation settings across multiple layers and attention heads to further examine Token-wise Attention Imbalance (TAI), and present three representative cases for demonstration.(Figure~\ref{case1}, Figure~\ref{case2}, and Figure~\ref{case3}) for illustration. Our key observations are summarized as follows.

\begin{enumerate}
    \item \textbf{TAI spans the entire generation sequence rather than only the final output tokens.}
    Visualizations across multiple layers and attention heads show that TAI does not appear solely within the output token sequence; instead, it emerges over large portions of the entire sequence and is particularly dense in the visual-token region.

    \item \textbf{TAI is more concentrated and pronounced in specific attention heads.}
    Different heads demonstrate markedly different attention patterns: certain heads (e.g., head 16) show stronger striped or slanted regularities, while others exhibit much weaker effects. This indicates that TAI is \emph{not uniformly distributed} across heads but manifests as \emph{head-specific biases}.

    \item \textbf{The strength of TAI varies structurally across layers.}
    From lower layers (Layer 0) to middle layers (Layer 15) and higher layers (Layer 31), attention patterns become increasingly regular and biased toward fixed token regions. This reveals that internal biases are progressively amplified or consolidated with depth, indicating a \emph{layer-wise accumulation or reinforcement} of TAI.

    \item \textbf{Averaged attention reveals global biases more clearly than individual heads.}
    Compared with per-head visualizations, the mean attention maps exhibit a more coherent slanted structure, showing that although individual heads behave differently, multi-head attention collectively forms a consistent sequence-level bias. This implies that TAI arises from a \emph{multi-head synergistic effect} rather than the behavior of isolated heads.
\end{enumerate}

\end{document}